\pdfoutput=1 
\documentclass[11pt,a4paper]{article}   
\usepackage[hyperref]{emnlp2020}
\usepackage{times}
\usepackage{latexsym}

\usepackage{microtype}

\aclfinalcopy 

\usepackage{algorithm,algorithmicx,algpseudocode}
\usepackage{url}
\usepackage{xcolor}
\usepackage[font={footnotesize}]{caption}
\usepackage{times}  
\usepackage{helvet} 
\usepackage{courier}  
\usepackage{booktabs}       
\usepackage{amsfonts}       
\usepackage{nicefrac} 
\usepackage{latexsym}
\usepackage{multirow}
\usepackage{booktabs}
\usepackage{amsmath}
\usepackage{booktabs}
\usepackage{subfig}
\usepackage{graphicx}  
\usepackage{xcolor}
\usepackage{colortbl}
\usepackage[normalem]{ulem}
\usepackage{CJKutf8}

\definecolor{myblue}{rgb}{0.54, 0.81, 0.94}
\definecolor{mygreen}{rgb}{0.64, 0.76, 0.68}
\definecolor{myyellow}{rgb}{0.98, 0.94, 0.75}
\definecolor{mygreen}{rgb}{0.68, 0.85, 0.9}
\definecolor{mypink}{rgb}{0.99, 0.87, 0.9}
\definecolor{myblue}{rgb}{0.82, 0.94, 0.75}	
\definecolor{brinkpink}{rgb}{0.98, 0.38, 0.5}
\definecolor{cadetgrey}{rgb}{0.5, 0.5, 0.5}

\definecolor{myblue2}{rgb}{0.36, 0.54, 0.66}

\definecolor{myorange}{rgb}{1.0, 0.49, 0.0}

\newcommand*\samethanks[1][\value{footnote}]{\footnotemark[#1]}

\title{\texttt{RethinkCWS}: Is \texttt{C}hinese \texttt{W}ord \texttt{S}egmentation a Solved Task?}

\author{First Author \\
  Affiliation / Address line 1 \\
  Affiliation / Address line 2 \\
  Affiliation / Address line 3 \\
  \texttt{email@domain} \\\And
  Second Author \\
  Affiliation / Address line 1 \\
  Affiliation / Address line 2 \\
  Affiliation / Address line 3 \\
  \texttt{email@domain} \\}

\date{}

\author{Jinlan Fu$\dag$ \thanks{\ \  These two authors contributed equally.}, \quad Pengfei Liu$\sharp$ \samethanks, \quad Qi Zhang$\dag$, \quad Xuanjing Huang$\dag$ \\
   $\dag$ School of Computer Science, Shanghai Key Laboratory \\
   of Intelligent Information Processing, Fudan University  \\
   $\sharp$Carnegie Mellon University \\
  \texttt{\{fujl16,qz,xjhuang\}@fudan.edu.cn}, \texttt{pliu3@cs.cmu.edu}}

\begin{document}
\maketitle
\begin{abstract}
The performance of the Chinese Word Segmentation (CWS) systems has gradually reached a plateau with the rapid development of deep neural networks, especially the successful use of large pre-trained models.
In this paper, we take stock of what we have achieved and rethink what's left in the CWS task.
Methodologically, we propose a fine-grained evaluation for existing CWS systems, which not only allows us to \emph{diagnose} the strengths and weaknesses of existing models (under the in-dataset setting), but enables us to \emph{quantify} the discrepancy between different criterion and alleviate the negative transfer problem when doing multi-criteria learning.
Strategically, despite not aiming to propose a novel model in this paper, our comprehensive experiments on eight models and seven datasets, as well as thorough analysis, could search for some promising direction for future research.
We make all codes publicly available and release an interface that can quickly evaluate and diagnose user's models: \url{https://github.com/neulab/InterpretEval}.




\end{abstract}

\begin{CJK*}{UTF8}{gbsn}
\section{Introduction}
Chinese word segmentation (CWS), as a crucial first step in Chinese language processing, has drawn a large body of research \cite{sproat1990statistical,xue2003chinese,huang2007rethinking,liu2014domain}.
Recent years have seen remarkable success in the use of deep neural networks on CWS \cite{zhou2017word,yang2017neural,ma2018state,yang2019subword,zheng2013deep,chen2015long,chen2015gated,cai2016neural,pei2014max}, and the large unsupervised pre-trained models drive the state-of-the-art results to a new level \cite{huang2019toward}.


However, the performance of CWS systems gradually reaches a plateau and the development of this field has slowed down. For example, the CWS systems on many existing datasets (e.g.~\texttt{msr}, \texttt{ctb}) have achieved $F1$-score higher than $97.0$ but with little further improvement. 
Naturally, a question would be raised: \emph{is CWS a solved task}?
When we rethink on what we have achieved so far, we find that there are still some important while rarely discussed unsolved questions for this task:


\textbf{Q1:}
Does current excellent performance (e.g. more than $98.0$ $F1$-score on the \texttt{msr} dataset) indicate a perfect CWS system, or are there still some limitations?
Existing CWS systems are mainly evaluated by a corpus-level metric.
The holistic measure fails to provide a fine-grained analysis. As a  result, we are not clear about  what the strengths and weaknesses of a specific model are.

To address this problem, we shift the traditional trend of holistic evaluation to fine-grained evaluation~\cite{fu2020interpretable}, in which the notion of the \textit{attribute} (i.e., \textit{word length}) has been introduced to describe a property of each word. Then test words will be partitioned into different buckets, in which we can observe the system's performances under different aspects based on word's attributes  (e.g.~\textit{long words} will obtain lower $F1$-score).



\textbf{Q2:}
Is there a one-size-fits-all system (i.e., best-performing systems on different datasets are the same)? If no, how can we make different choices of model architectures in different datasets?
Insights are still missing for how the choices of different datasets influence architecture design.

To answer this question, we make use of our proposed fine-grained evaluation methodology and present two types of diagnostic methods for existing CWS systems, which not only helps us to identify the strengths and weaknesses of current approaches but provides us with more insight about how different choices of datasets influence the model design.

\textbf{Q3:} Now that existing works show CWS systems can benefit from multi-criteria learning at the cost of negative transfer \cite{chen2017adversarial,qiu2019multi-criteria}, can we design a measure to quantify the discrepancies among different criteria and use it to instruct the multi-criteria learning process (i.e., alleviate negative transfer)?

\renewcommand\tabcolsep{1.8pt}
\begin{table}[t]
  \centering \footnotesize
    \begin{tabular}{ccccc}
    \toprule
    Settings & \multicolumn{3}{c}{Measures} & Application \\
     \cmidrule(lr){1-1} \cmidrule(lr){2-4} \cmidrule(lr){5-5}
     & \multirow{2}[1]{*}{Model} & Spearman & $S^{\rho}$ (Eq.~\ref{eq:m_rho})     &   \\
    In-   &   & Variance & $S^{\sigma}$ (Eq.~\ref{eq:m_sigma})   & Model Diagnosis \\
          \cmidrule(lr){2-4} 
    Dataset  & \multirow{2}[1]{*}{Data} & Sys-indep & $\alpha^{\mu}$ (Eq.~\ref{eq:alpha_mu})      & Sec.~\ref{sec:model-diagmosis} (Q1,Q2) \\
    &       & Sys-dep & $\alpha^{\rho}$ (Eq.~\ref{eq:alpha_rho})    &  \\
    \midrule
    Cross- & Model & Generali. & $\mathbf{U}$ (Eq.~\ref{eq:generali})    & \multirow{1}[2]{*}{Multi-Source Tra- } \\
    \cmidrule(lr){2-4} 
      Dataset    & Data  & Criterion & $\Psi$ (Eq.~\ref{equ:psi})      & nsfer Sec.~\ref{sec:order-matter} (Q3)\\
    \bottomrule
    \end{tabular}%
    \vspace{-6pt}
    \caption{An outline of our paper. \textit{``Generali.''}, \textit{``Sys''}, \textit{``indep''}, and \textit{``dep''} are the abbreviation for ``Generalization'', ``System'', ``independent'', and ``dependent'', respectively.  }
  \label{tab:outline}
\end{table}%

To answer this question, we extend the \textit{in-dataset} evaluation (i.e., a system is trained and tested on the same dataset) to the setting of \textit{cross-dataset}, in which a CWS model trained on one corpus would be evaluated on a range of out-of-domain corpora.
On the other hand, it's the above in-dataset analysis (in Q1 \& Q2) that helps us to design a measure to quantify the discrepancies of cross-dataset criterion.
Empirical results not only show that the measure, calculated solely based on statistics of two datasets, has a higher correlation with cross-dataset performances but also helps us avoid the negative transfer (i.e., selecting the useful parts of source domains as training sets and achieve better results based on fewer training samples)

Our contributes can be summarized as follows:
1) Instead of using a holistic metric, we  proposed an attribute-aided evaluation methodology for CWS systems. This allows us to diagnose the weakness of existing CWS systems (e.g., BERT-based models are not impeccable and limited in dealing with words with high label inconsistency).
2) We show that best-performing systems on different datasets are diverse. Based on some proposed quantified measures, we can make good choices of model architectures in different datasets.
3) We quantify the criterion discrepancy between different datasets, which can alleviate the negative transfer problem when performing multi-criteria learning for CWS.


\section{Preliminaries}

\subsection{Task Description}
Chinese word segmentation (CWS) was usually conceptualized as a character-based sequence labeling problem. 
Formally, let $X = \{x_1,x_2,\ldots, x_T\}$ be a sequence of characters, and $Y = \{y_1,y_2,\ldots,y_T\}$ be the output tags.
The goal of the task is to estimate the conditional probability:
$P(Y | X) = P(y_t|X,y_1,\cdots,y_{t-1})$.
Here, $y_t$ usually takes one value of $\{B, M, E, S\}$.

\subsection{Attribute-aided Evaluation Methodology} \label{sec:fine-grained}
\begin{figure}[!t]
\centering
\includegraphics[width=0.80\linewidth]{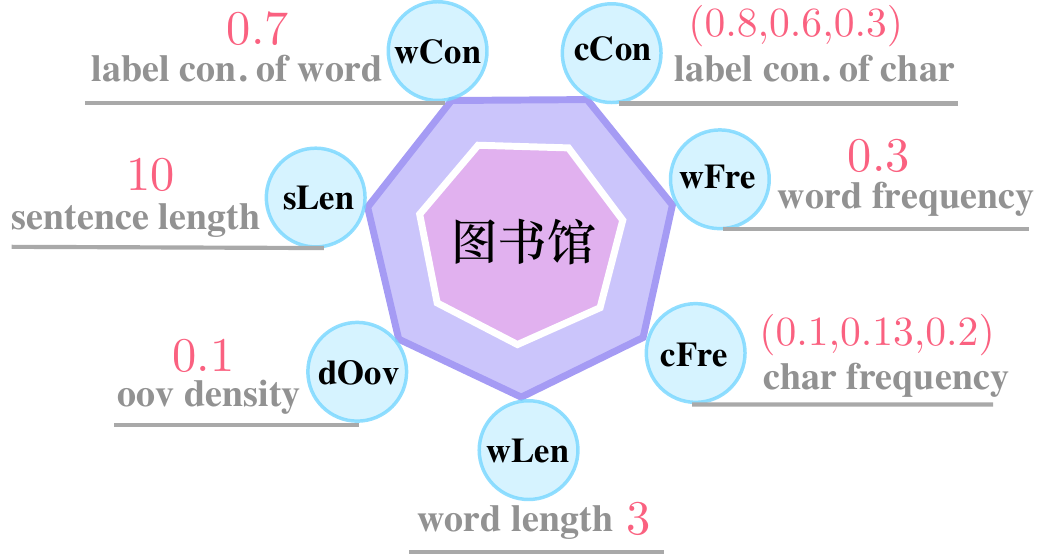}
\vspace{-6pt}
\caption{\footnotesize{The attribute definition of the word ``\texttt{图书馆 (library)}'' in the sentence: ``图书馆在节假日会关闭 (The \texttt{library}  is closed on holidays'', and its ground truth label is \texttt{BME}. The text in the circle is the abbreviation of the attribute name, and the text in gray and in pink is the full name and the attribute value, respectively. \textit{con.} in grey denotes \textit{consistency}. }} 

\label{fig:seattle}
\end{figure}

The standard metric of CWS is becoming hard to distinguish the state-of-the-art word segmentation systems \cite{qian2016a}.
Instead of evaluating CWS systems based on a holistic metric (F1 score), in this paper, we take a step towards the fine-grained evaluation of the current CWS systems by proposing an attribute-aided evaluation method.
Specifically, we first introduce the notion of \textit{attributes} to characterize the properties of the test words. Then, the test set will be divided into different subsets, and the overall performance could be broken down into several interpretable \textit{buckets}.
Below, we will introduce the \textit{seven} attributes that we have explored to depict the word in diverse aspects.  Fig.~\ref{fig:seattle} gives an example for the test word ``\texttt{图书馆}''.

\paragraph{Aspect-I: Intrinsic nature}
We can characterize a word based on its (or the sentence it belongs to) constitute features. Here, we define three attributes: \underline{word length (\texttt{wLen})}; \underline{sentence length (\texttt{sLen})}; \underline{OOV density (\texttt{oDen})}: the number of words outside the training set in a sentence divided by sentence length.

\paragraph{Aspect-II: Familiarity}
We introduce a notion of familiarity to quantify the degree to which a test word (or its constituents) has been seen in the training set. 
Specifically, the familiarity of a word can be calculated based on its \textit{frequency} in the training set. 
For example, in Fig.~\ref{fig:seattle}, if the frequency in the training set of the test word \texttt{图书馆} (library) is \texttt{0.3}, the attribute of word frequency of \texttt{图书馆} will be \texttt{0.3}.
In this paper, we consider two kinds of familiarity:  \underline{word frequency (\texttt{wFre})};  \underline{character frequency (\texttt{cFre})}.

\paragraph{Aspect-III: Label consistency}
In this paper, we attempt to design a measure that can quantify the degree of label consistency phenomenon \cite{fu2020rethinking,gong2017multi,luo2016empirical,chen2017adversarial} for each test word (or character). Here, we investigate two attributes for label consistency:  \underline{label consistency of word (\texttt{wCon})}; \underline{label consistency of character (\texttt{cCon})}. Following, we give the definition of label consistency of word, and the label consistency of character can be defined in a similar way.
Specifically, we refer to $w_i^k$ as a test word with label k, whose label consistency $\psi(w_i^{k})$ is defined as:
\vspace{-5pt}
\begin{equation}
    \psi(w_{i}^{k}, D^{tr}) =
  \begin{cases}
  0 & \mbox{$|w_{i}^{tr}|=0$ }\\
  \frac{|w_{i}^{tr,k}|}{|w_{i}^{tr}|}
   &\mbox{otherwise}
  \end{cases} \label{eq:def-rho}
\end{equation} 
\vspace{-15pt}

where $|w_{i}^{tr,k}|$ represents the occurrence of word $w_i$ with label $k$ in the training set, and 
$D^{tr}$ is the training set. 
For example, in Fig.~\ref{fig:seattle}, in the training set, ``\texttt{图书馆} (library)'' is labeled as  \texttt{BME} 7 times, and \texttt{BMM} 3 times, so $\psi$ (`` \texttt{图书馆$^{BME}$}'') $=7/10 = 0.7$, and $\psi$ (``\texttt{图书馆$^{BMM}$}'') $=3/10 = 0.3$ . 

\vspace{-5pt}

\section{Investigation on In-dataset Setting}
\subsection{Setup} \label{sec:indomain-setup}
This section focuses on the \textit{in-dataset} setting, in which each CWS model will be trained and test on the same dataset.

\paragraph{Datasets}
We choose seven mainstream datasets from  SIGHAN2005 \footnote{http://sighan.cs.uchicago.edu/bakeoff2005/} and SIGHAN2008 \footnote{https://www.aclweb.org/mirror/ijcnlp08/sighan6/chinese bakeoff.htm}, in which \texttt{cityu} and \texttt{ckip} are traditional Chinese, while \texttt{msr}, \texttt{pku}, \texttt{ctb}, \texttt{ncc} and \texttt{sxu} are simplified Chinese. We map traditional Chinese characters to simplified Chinese in our experiment. The details of the seven datasets used in this study are described in \citet{chen2017adversarial}.

\paragraph{Models}
We choose typical instances as analytical objects, which vary in terms of the following aspects:  
1) {character encoders}:
\texttt{ELMo} \citep{peters2018deep}, 
\texttt{BERT} \citep{devlin2018bert}; 
2) {bigram encoder}: \texttt{Word2Vec} \citep{mikolov2013efficient}, averaging the embedding of two contiguous characters; 
3) sentence encoders: \texttt{LSTM} \citep{hochreiter1997long}, \texttt{CNN} \citep{kalchbrenner2014convolutional};
4) {decoders}: \texttt{MLP}, \texttt{CRF} \citep{lample2016neural,collobert2011natural}.
The name of combination of models in in a detailed setting in Tab.\ref{tab:holistic}.

\subsection{Measures} \label{sec:indomain-measure}
Here, we refer to $M = \{m_1,\cdots, m_{N_m}\}$ as a set of \textbf{models} and $P = \{p_1,\cdots,p_{N_p}\}$ as a set of \textbf{attributes}.
As described above, the test set could be split into different \textbf{buckets} $B = \{B^j_{1}, \cdots, B^j_{N_b} \}$ based on an attribute $p_j$.
We introduce a performance table ${\mathbf{V}} \in \mathbb{R}^{N_m\times N_p \times N_b}$, in which $\mathbf{V}_{ijk}$ represents the performance of $i$-th model on the $k$-th sub-test set (bucket) generated by $j$-th attribute.

\paragraph{Model-wise}
The model-wise measure aims to investigate
whether and how the attributes influence the performance of models with different choices of neural components.
Formally, we characterize how the $j$-th attribute influences the $i$-th model based on two statistical variables: Spearman's rank correlation coefficient $\mathrm{Spear}$ \citep{mukaka2012guide} and standard deviation $\mathrm{Std}$, which can be defined as:

\vspace{-10pt}
\setlength\abovedisplayskip{-2pt}
\begin{align}
    \mathbf{S}^{\rho}_{i,j} &= \mathrm{Spear}(\mathbf{V}[i,j:], R_{j}), \label{eq:m_rho} \\
    \mathbf{S}^{\sigma}_{i,j} &= \mathrm{Std}(\mathbf{V}[i,j:]), \label{eq:m_sigma}
\end{align}
\vspace{-20pt}

where $R_{j}$ is the rank values of buckets based on $j$-th attribute. 
Intuitively, $\mathbf{S}^{\rho}_{i,j}$ reflects the degree to which the $i$-th model positively (negatively) correlates with $j$-th attribute while $\mathbf{S}^{\sigma}_{i,j}$ indicates the degree to which this attribute influences the model.

\renewcommand{\arraystretch}{1.2}
\begin{table*}[!htb]
  \centering  \small
    \begin{tabular}{ccccccccccccccccccc}
    \toprule
    \multirow{2}[4]{*}{\textbf{Model}} & \multicolumn{4}{c}{\textbf{Character}} & \multicolumn{3}{c}{\textbf{Bigram}} & \multicolumn{2}{c}{\textbf{SenEnc.}} & \multicolumn{2}{c}{\textbf{Dec.}} & \multicolumn{7}{c}{\textbf{Holistic Evaluation (Overall F1)}} \\
    \cmidrule(lr){2-5} \cmidrule(lr){6-8} \cmidrule(lr){9-10} \cmidrule(lr){11-12} 
& \multicolumn{1}{c}{\rotatebox{90}{rand}}  & \multicolumn{1}{c}{\rotatebox{90}{w2v} }  & \multicolumn{1}{c}{\rotatebox{90}{elmo} } & \multicolumn{1}{c}{\rotatebox{90}{bert} } & \multicolumn{1}{c}{\rotatebox{90}{none}}  & \multicolumn{1}{c}{\rotatebox{90}{avg} }  & \multicolumn{1}{c}{\rotatebox{90}{w2v}  } & \multicolumn{1}{c}{\rotatebox{90}{lstm}  }& \multicolumn{1}{c}{\rotatebox{90}{cnn} }  & \multicolumn{1}{c}{\rotatebox{90}{crf} }  & \multicolumn{1}{c}{\rotatebox{90}{mlp} }  & \textbf{msr}     & \textbf{pku}   & \textbf{ctb}   & \textbf{ckip}  & \textbf{cityu} & \textbf{ncc}   & \textbf{sxu}\\
    \midrule
    CrandBavgLstmCrf & $\surd$     &       &       &       &       & $\surd$     &       & $\surd$     &       & $\surd$     &       & 96.21   & \textbf{94.22} & \textbf{95.32} & \textbf{92.81} & 93.54 & 92.01 & 94.87 \\   
    Cw2vBavgLstmCrf &       & $\surd$     &       &       &       & $\surd$     &       & $\surd$     &       & $\surd$     &       & 96.46  & 94.10  & 95.08 & 92.81 & 93.67 & 92.04 & 94.71 \\    
    Cw2vBavgLstmMlp &       & $\surd$     &       &       &       & $\surd$     &       & $\surd$     &       &       & $\surd$     & 96.41    & 92.74 & 94.09 & 91.40  & 93.25  & 92.00    & 93.16 \\    
    Cw2vBavgCnnCrf &       & $\surd$     &       &       &       & $\surd$     &       &       & $\surd$     & $\surd$     &       & 96.48  & 93.99 & 94.72 & 92.73 & \textbf{93.72} & \textbf{92.64} & 94.36 \\

    Cw2vBw2vLstmCrf &       & $\surd$     &       &       &       &       & $\surd$     & $\surd$     &       & $\surd$     &       & \textbf{96.66}  & 94.19 & 95.14 & 92.46 & 93.70  & 92.24 & \textbf{94.97} \\
    
    \midrule
    CelmBnonLstmMlp &       &       & $\surd$     &       & $\surd$     &       &       & $\surd$     &       &       & $\surd$     & 96.23  & 95.33 & 96.77 & 94.83 & 96.44 & 93.21 & 96.47 \\
    CbertBnonLstmMlp &       &       &       & $\surd$     & $\surd$     &       &       & $\surd$     &       &       & $\surd$    & 98.19  & 96.47 & \textbf{97.68} & \textbf{96.23} & 97.09 & 95.77 & 97.49 \\
    CbertBw2vLstmMlp &       & $\surd$     &       & $\surd$     &       &       & $\surd$     & $\surd$     &       &       & $\surd$     & \textbf{98.20}  & 96.52 & 97.65 & 96.18 & 97.07 & \textbf{95.78} & \textbf{97.51} \\
    \citet{huang2019toward} &       &      &       &      &       &       &      &      &       &       &     & 97.90  & \textbf{96.60} & 97.60 &---  & \textbf{97.60} &---   & 97.30 \\
    \bottomrule
    \end{tabular}%
    \vspace{-6pt}
    \caption{Neural CWS systems with different architectures and pre-trained knowledge studied in this paper. We exclude systems based on joint training to make a fair comparison in the in-dataset setting. For the model name, ``\texttt{C}" refers to ``\texttt{Character}" and ``\texttt{B}" refers to ``\texttt{Bigram}". Intuitively, the models are named based on their constituents. For example, \textit{Cw2vBw2vLstmCrf} denotes a model's character and the bigram feature is initialized by pre-trained embeddings using Word2Vec, and sentence encoder, as well as the decoder, are LSTM and CRF, respectively. We perform a Friedman test at p = 0.05 on 
   model- (row-) wise  and data- (column-)wise. The testing results are $p(\mathrm{model-wise}) = 2.26 \times 10^{-6} <0.05$ and $p(\mathrm{data-wise}) = 8.42 \times 10^{-8}$. Therefore, the results of model-wise and data-wise have passed the significance testing.}
  \label{tab:holistic}%
\end{table*}%

\paragraph{Dataset-wise}

The dataset-wise measures aim to characterize a dataset with different attributes quantitatively.
We utilize two types of measures to build the connection between datasets and attributes:
system-independent measure $\alpha^{\mu}$, and system-dependent measures $\alpha^{\rho}$ and $\alpha^{\sigma}$.

1) \textit{system-independent measure} reflects intrinsic statistics of the datasets, such as the average word length of the whole dataset. It can be formally defined as: 
\setlength\abovedisplayskip{-1pt}
\begin{align}
\alpha^{\mu}_j = \frac{1}{N_w} \sum_i^{N_w} \mathrm{Attr}(w_i, j)\label{eq:alpha_mu},
\end{align}
where $N_w$ is the number of test words and $\mathrm{Attr}(w_i, j)$ is the value of attribute $j$ for word $w_i$.

2) \textit{system-dependent measures} quantify the degree to which each attribute influences the CWS system on a given dataset.
For example, ``does the attribute \texttt{word length} matter for the CWS system trained on \texttt{pku} dataset?".
To achieve this, we design the following measures:
\begin{align}
    \alpha^{\rho}_j &= \frac{1}{N_m} \sum_i^{N_m} |\mathbf{S}^{\rho}_{i,j}|, 
    \label{eq:alpha_rho}
\end{align}
\vspace{-15pt}

where $N_m$ is the number of evaluated models. Intuitively, a higher absolute value of $\alpha^{\rho}_j \in [-1,1]$ suggests that attribute $j$ is a crucial factor, greatly influencing the performance of CWS systems.
For example, if $\alpha^{\rho}_{wLen} = 0.95$, it means \texttt{word length} is a major factor that influences the CWS performance.

\begin{figure}[!htb]
    \centering
    \subfloat[$\alpha^{\mu}$]{
    \includegraphics[width=0.39\linewidth]{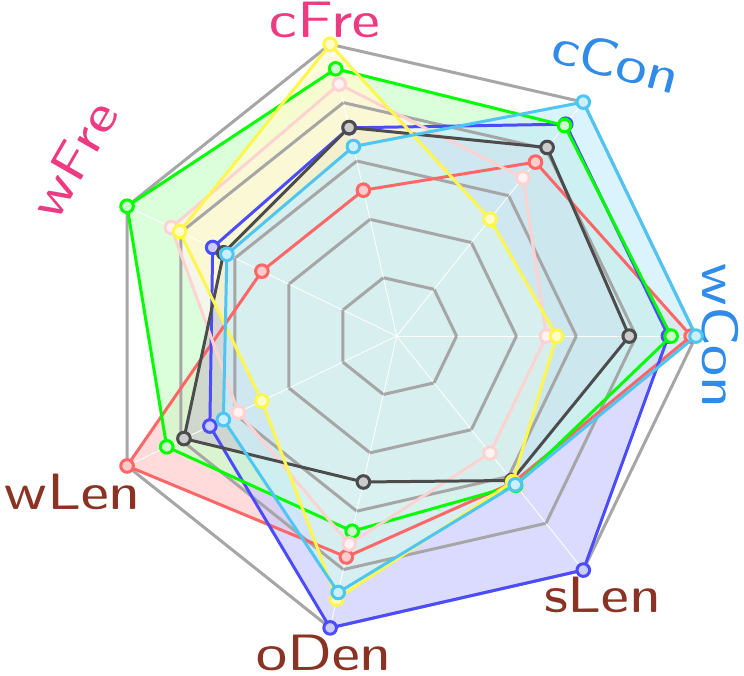} 
    }  \hspace{0.4em}  
    \subfloat[$\alpha^{\rho}$]{
    \includegraphics[width=0.52\linewidth]{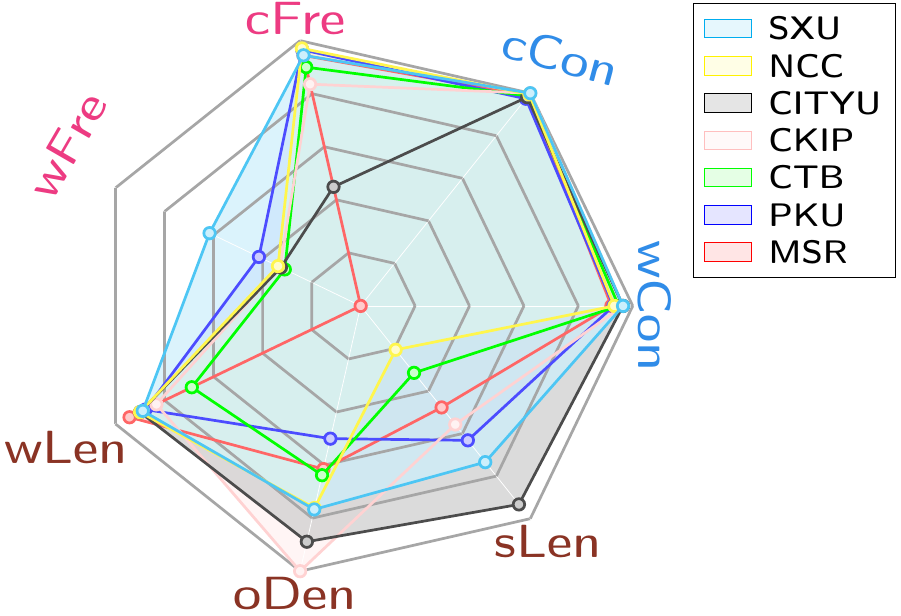} 
    }
    \vspace{-6pt}
    \caption{Illustration of dataset biases characterized by task-independent measure $\alpha^{\mu}$ and task-dependent measures $\alpha^{\rho}$. We normalize  $\alpha^{\mu}$ on each attribute by divide the maximum  $\alpha^{\mu}$ on six datasets, and $\alpha^{\rho} \in [0,1]$.}
    \label{fig:spider}
\end{figure}
\vspace{-3pt}

\subsection{Analysis of Holistic Evaluation}
Before giving a fine-grained analysis, we present the holistic results of different models on different datasets.
As shown in Tab.~\ref{tab:holistic},  we can observe that \textbf{there is no one-size-fits-all model: best-performing systems on different datasets frequently consist of diverse components}. This naturally raises a question: how can pick up appropriate models for different datasets?

\subsection{Analysis of Dataset Biases}
Before the analysis, we conduct a statistical significance test with the Friedman test~\cite{zimmerman1993relative} at $p=0.05$, to examine whether the performance of different buckets partitioned by an attribute is significantly different for a given dataset. The results are shown in the Appendix. We find that the performance of different buckets partitioned by an attribute is significantly different ($p<0.05$), which holds for all the datasets.

1) \textbf{\textit{Label consistency} and \textit{word length} have a more consistent impact on CWS performance.}
The common parts of the radar charts Fig.~\ref{fig:spider} (b) illustrate that
no matter which datasets are, 
label consistency attributes (\texttt{wCon}, \texttt{cCon}) and word length (\texttt{wLen}) are  highly correlated with CWS performance (higher $\alpha^{\rho}$).
This suggests that the learning difficulty of CWS systems is commonly influenced by label consistency and word length.

2) \textbf{\textit{Frequency} and \textit{sentence length} matters but are minor factors}
The outliers in the radar chart (Fig.~\ref{fig:spider} (b)) show the peculiarities of different corpora.
On attributes: \texttt{sLen}, \texttt{wFre}, \texttt{oDen}, the extent to which different datasets are affected varies greatly.
For example, the dataset \texttt{ckip} is distinctive with the highest value of $\alpha_{oDen}^{\rho}$, which can explain why character pre-training shows no advantage while the CRF layer contributes a lot. 


\subsection{Analysis of Model Biases}
Similar to the above section, we perform the Friedman test at $p=0.05$. 
We give detailed significance testing results  in the Appendix. 
Tab.~\ref{tab:spear-std} gives an illustration of model biases characterized by measures $\mathbf{S}^{\rho}_{i,j}$ and  $\mathbf{S}^{\sigma}_{i,j}$.
The values in grey denote the given model on the specific attribute does not pass the significance test ($p\geq 0.05$).
Below, we will highlight some observations.



\paragraph{ELMo-based Models can make better use of the context information that long sentences carry.}
Regarding the attribute of \texttt{sLen} (sentence length), two models \textit{CelmBnonLstmMlp} and \textit{CbertBnonLstmMlp} pass the significance test. 
Additionally, we observe only ELMo (\textit{CelmBnonLstmMlp}) shows a strong positive correlation with sentence length, referring to Tab.~\ref{tab:spear-std}.


\paragraph{Contextualized models could reduce the negative effect of OOV density and remedy the deficiency of MLP decoder.}
a) The performances of non-contextualized models (i.e. word2vec) strongly correlate with the \texttt{oDen} (density of OOV words) attribute.
When equipped with BERT or ELMo, the model still could provide each OOV word with a meaningful representation on the fly based on its context. 
b) We observe that the model \textit{Cw2vBavgLstmMlp} is strong correlated with \texttt{wCon} and \texttt{wLen} with highest values of $\mathbf{S}^{\sigma}$ (referring to Tab.~\ref{tab:spear-std} with bolded value), suggesting that models with MLP layer are unstable when generalizing to the hard cases (words with lower value of  \texttt{wCon} and higher value of \texttt{wLen}). However, once augmented with contextualized models, systems with MLP decoder also work well.

\renewcommand\tabcolsep{0.7pt}
\begin{table}[!htb]
  \centering \scriptsize
    \begin{tabular}{lcrrrrrrrrrrrrrr}
    \toprule
    & & \multicolumn{7}{c}{\textbf{Spearmanr }}                & \multicolumn{7}{c}{\textbf{Standard Deviation }} \\
    \cmidrule(lr){3-9} \cmidrule(lr){10-16}
          \textbf{Model} &  \textbf{F1} &
          \multicolumn{1}{c}{\rotatebox{90}{wCon}} & \multicolumn{1}{c}{\rotatebox{90}{cCon}} &  \multicolumn{1}{c}{\rotatebox{90}{cFre}} & \multicolumn{1}{c}{\rotatebox{90}{wFre}} &
          \multicolumn{1}{c}{\rotatebox{90}{wLen}} &          
          \multicolumn{1}{c}{\rotatebox{90}{oDen}} & 
          \multicolumn{1}{c}{\rotatebox{90}{sLen}} &   
          \multicolumn{1}{c}{\rotatebox{90}{wCon}} & \multicolumn{1}{c}{\rotatebox{90}{cCon}} &  \multicolumn{1}{c}{\rotatebox{90}{cFre}} & \multicolumn{1}{c}{\rotatebox{90}{wFre}} &
          \multicolumn{1}{c}{\rotatebox{90}{wLen}} &          
          \multicolumn{1}{c}{\rotatebox{90}{oDen}} & 
          \multicolumn{1}{c}{\rotatebox{90}{sLen}}
          \\
    \midrule
    CrandBavgLstmCrf & 94.14 & 92  & 99  & 88  & 33  & -85 & \textcolor{cyan}{-82}   & \textcolor{cadetgrey}{20}  & 13  & 9.3   & 2.4   & 6.4   & 13  & 1.6   & \textcolor{cadetgrey}{0.6}  \\
    Cw2vBavgLstmCrf & 94.12 & 93  & 99  & 91  & 33  & -85  & \textcolor{cyan}{-86}  & \textcolor{cadetgrey}{18}  & 13  & 10  & 2.4   & 7.3   & 13  & 1.8   & \textcolor{cadetgrey}{0.6}  \\
    Cw2vBavgLstmMlp & 93.29 & 95  & 98  & 93  & 37  & -86  & \textcolor{cyan}{-76}  & \textcolor{cadetgrey}{8.9}   & \textcolor{cyan}{19}  & {11}  & {3.1}   & {7.9}   & \textcolor{cyan}{15}  & {2.7}   & \textcolor{cadetgrey}{1.2}  \\
    Cw2vBavgCnnCrf & 94.09 & 96  & 99  & 92  & 35  & -86  & \textcolor{cyan}{-73}  & \textcolor{cadetgrey}{17}  & 15  & 9.4   & 2.5   & 7.0   & 14  & 1.5   & \textcolor{cadetgrey}{0.7}  \\
    Cw2vBw2vLstmCrf & 94.20 & 93  & 99  & 90  & 33  & -89  & \textcolor{cyan}{-85}  & \textcolor{cadetgrey}{28}  & 13  & 10  & 2.4   & 7.5   & 13  & 1.9   & \textcolor{cadetgrey}{0.6}  \\
    \midrule
    CelmBnonLstmMlp & 95.61 & 95  & 98  & 78  & 31  & -82 & \textcolor{cyan}{-44}  & \textcolor{myorange}{73}  & \textcolor{cyan}{9.0}   & 5.1   & 1.4   & 4.5   & \textcolor{cyan}{8.2}   & 1.5   & 0.5  \\
    CbertBnonLstmMlp & 96.99 & 96  & 98  & 74  & 34  & -88  & \textcolor{cyan}{-30} & \textcolor{myorange}{39}  & \textcolor{cyan}{6.2}   & 3.7   & 1.0   & 2.8   & \textcolor{cyan}{5.8}   & 1.2   & 0.3  \\
    CbertBw2vLstmMlp & 97.00 & 96  & 99  & 77  & 30  & -86  & \textcolor{cyan}{-29} & \textcolor{cadetgrey}{37}  & \textcolor{cyan}{6.3}   & 3.9   & 1.0   & 2.8   & \textcolor{cyan}{5.8}   & 1.2   & \textcolor{cadetgrey}{0.3}  \\
    \bottomrule
    \end{tabular}%
    \vspace{-6pt}
  \caption{Illustration of model biases characterized by model-wise measure (Percentage) $ \mathbf{S}^{\rho}_{i,j}$ and  $\mathbf{S}^{\sigma}_{i,j}$. Here, we average the F1, $ \mathbf{S}^{\rho}_{i,j}$ and $\mathbf{S}^{\sigma}_{i,j}$ on seven datasets. The values in gray denotes the given model on the specific attribute does not pass the significance test ($p\geq 0.05$). The values in \textcolor{myorange}{orange} and in \textcolor{cyan}{blue} support observation 1 and observation 2, respectively. 
}
   \label{tab:spear-std}%
\end{table}%
\vspace{-10pt}

\renewcommand\tabcolsep{0.75pt}
\renewcommand\arraystretch{0.68}  
\begin{table*}[!htb]
  \centering \tiny
    \begin{tabular}{ccccccc ccccccc ccccccc ccccccc ccccccc ccccccc ccccccc cccccccccccccccccccccccccccccccccccccccccccccccccc}
    \toprule
          & \multicolumn{14}{c}{msr }                           & \multicolumn{14}{c}{pku}                            & \multicolumn{14}{c}{ctb}                            & \multicolumn{14}{c}{ckip}  &
          \multicolumn{14}{c}{cityu}  &
          \multicolumn{14}{c}{ncc}  &
          \multicolumn{14}{c}{sxu}\\
    \midrule
    \multicolumn{1}{l}{} & &&&&&&& \multicolumn{1}{l}{\rotatebox{90}{wCon}} & \multicolumn{1}{l}{\rotatebox{90}{cCon}} & \multicolumn{1}{l}{\rotatebox{90}{cFre}} & \multicolumn{1}{l}{\rotatebox{90}{wFre}} & \multicolumn{1}{l}{\rotatebox{90}{wLen}} & \multicolumn{1}{l}{\rotatebox{90}{oDen}} & \multicolumn{1}{l}{\rotatebox{90}{sLen}} &
    &&&&&&&
    \multicolumn{1}{l}{\rotatebox{90}{wCon}} & \multicolumn{1}{l}{\rotatebox{90}{cCon}} & \multicolumn{1}{l}{\rotatebox{90}{cFre}} & \multicolumn{1}{l}{\rotatebox{90}{wFre}} & \multicolumn{1}{l}{\rotatebox{90}{wLen}} & \multicolumn{1}{l}{\rotatebox{90}{oDen}} & \multicolumn{1}{l}{\rotatebox{90}{sLen}} &
    &&&&&&&
    \multicolumn{1}{l}{\rotatebox{90}{wCon}} & \multicolumn{1}{l}{\rotatebox{90}{cCon}} & \multicolumn{1}{l}{\rotatebox{90}{cFre}} & \multicolumn{1}{l}{\rotatebox{90}{wFre}} & \multicolumn{1}{l}{\rotatebox{90}{wLen}} & \multicolumn{1}{l}{\rotatebox{90}{oDen}} & \multicolumn{1}{l}{\rotatebox{90}{sLen}} &
    &&&&&&&
    \multicolumn{1}{l}{\rotatebox{90}{wCon}} & \multicolumn{1}{l}{\rotatebox{90}{cCon}} & \multicolumn{1}{l}{\rotatebox{90}{cFre}} & \multicolumn{1}{l}{\rotatebox{90}{wFre}} & \multicolumn{1}{l}{\rotatebox{90}{wLen}} & \multicolumn{1}{l}{\rotatebox{90}{oDen}} & \multicolumn{1}{l}{\rotatebox{90}{sLen}} &
    &&&&&&&
    \multicolumn{1}{l}{\rotatebox{90}{wCon}} & \multicolumn{1}{l}{\rotatebox{90}{cCon}} & \multicolumn{1}{l}{\rotatebox{90}{cFre}} & \multicolumn{1}{l}{\rotatebox{90}{wFre}} & \multicolumn{1}{l}{\rotatebox{90}{wLen}} & \multicolumn{1}{l}{\rotatebox{90}{oDen}} & \multicolumn{1}{l}{\rotatebox{90}{sLen}} &
    &&&&&&&
    \multicolumn{1}{l}{\rotatebox{90}{wCon}} & \multicolumn{1}{l}{\rotatebox{90}{cCon}} & \multicolumn{1}{l}{\rotatebox{90}{cFre}} & \multicolumn{1}{l}{\rotatebox{90}{wFre}} & \multicolumn{1}{l}{\rotatebox{90}{wLen}} & \multicolumn{1}{l}{\rotatebox{90}{oDen}} & \multicolumn{1}{l}{\rotatebox{90}{sLen}} &
    &&&&&&&
    \multicolumn{1}{l}{\rotatebox{90}{wCon}} & \multicolumn{1}{l}{\rotatebox{90}{cCon}} & \multicolumn{1}{l}{\rotatebox{90}{cFre}} & \multicolumn{1}{l}{\rotatebox{90}{wFre}} & \multicolumn{1}{l}{\rotatebox{90}{wLen}} & \multicolumn{1}{l}{\rotatebox{90}{oDen}} & \multicolumn{1}{l}{\rotatebox{90}{sLen}} 
    \\
    \midrule
     \multicolumn{1}{l}{Overall F1} & 
     \multicolumn{14}{c}{A: 98.19} &
     \multicolumn{14}{c}{A: 96.47} &
     \multicolumn{14}{c}{A: 97.68}  &
     \multicolumn{14}{c}{A: 96.23} &
     \multicolumn{14}{c}{A: 97.09}&
     \multicolumn{14}{c}{A: 95.77}&
     \multicolumn{14}{c}{A: 97.49}\\
    \cmidrule(r){1-1}\cmidrule(lr){2-15}\cmidrule(lr){16-29}\cmidrule(lr){30-43}\cmidrule(lr){44-57}\cmidrule(lr){58-71}\cmidrule(lr){72-85}\cmidrule(lr){86-99}

    \multirow{3}[2]{*}{A:C\textcolor{myblue2}{bert}BnonLstmMlp} & \multicolumn{14}{c}{\multirow{3}[2]{*}{\includegraphics[scale=0.26]{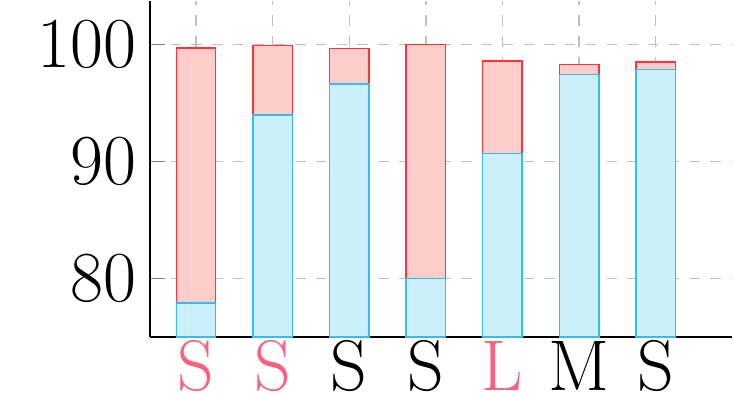}}}              & \multicolumn{14}{c}{\multirow{3}[2]{*}{\includegraphics[scale=0.26]{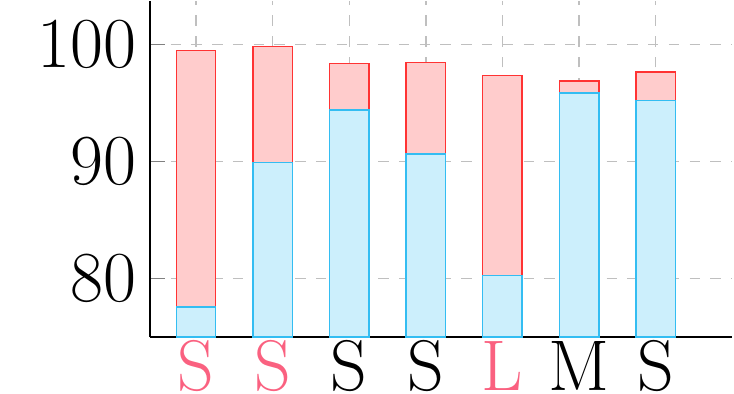}}}              & \multicolumn{14}{c}{\multirow{3}[2]{*}{\includegraphics[scale=0.26]{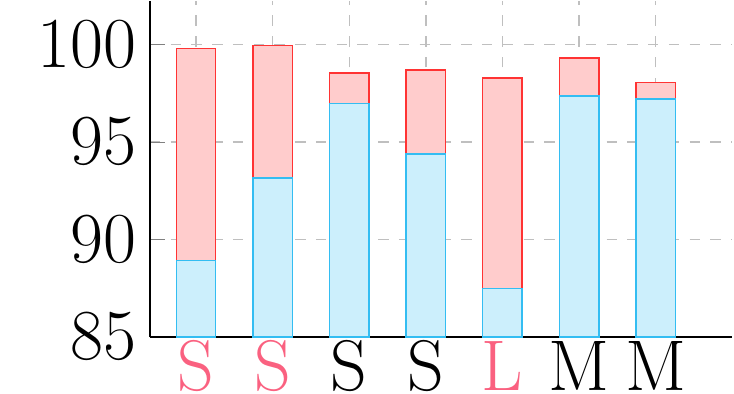}}}              & \multicolumn{14}{c}{\multirow{3}[2]{*}{\includegraphics[scale=0.26]{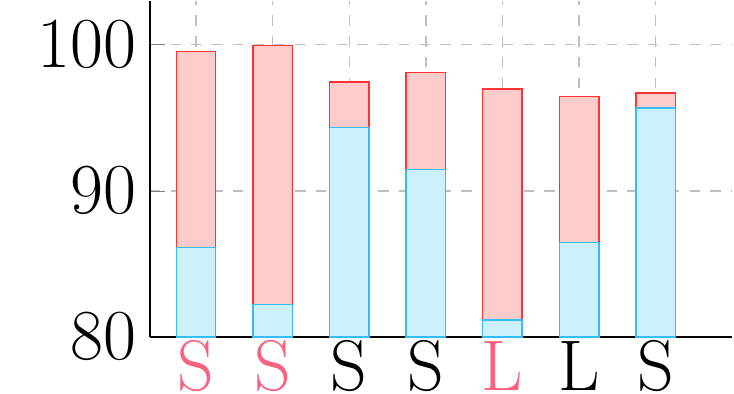}}} &
    \multicolumn{14}{c}{\multirow{3}[2]{*}{\includegraphics[scale=0.26]{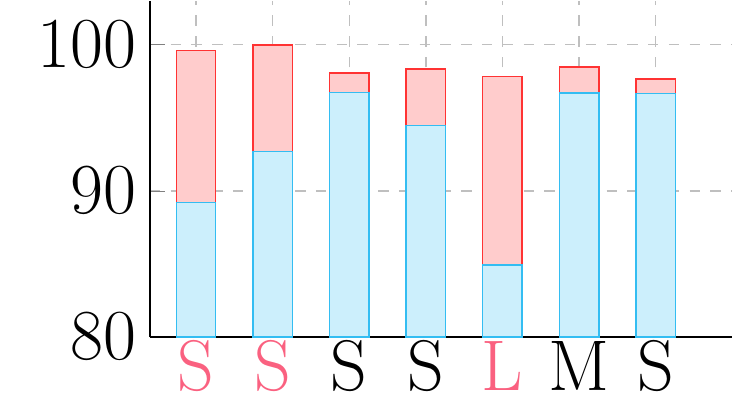}}} &
    \multicolumn{14}{c}{\multirow{3}[2]{*}{\includegraphics[scale=0.26]{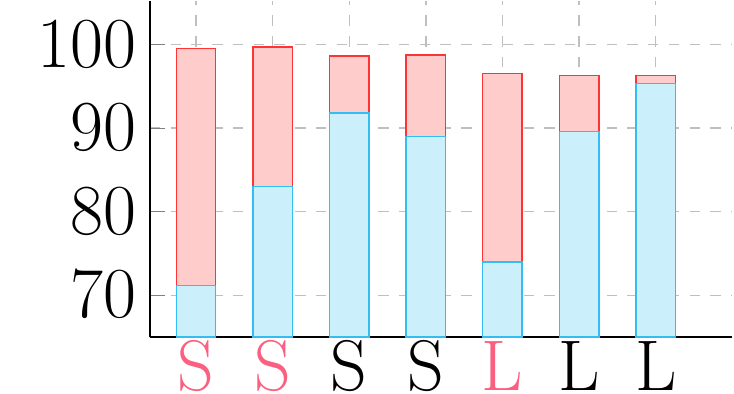}}} &
    \multicolumn{14}{c}{\multirow{3}[2]{*}{\includegraphics[scale=0.26]{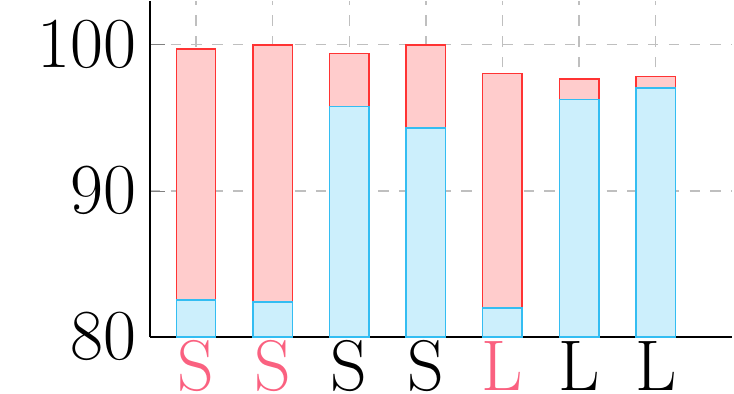}}} 
    \\  
    \\ \\ \\ 
    \multicolumn{1}{c}{\textcolor{brinkpink}{Self-diagnosis}} & 
    \\ \\  
    
    \midrule
     \multicolumn{1}{l}{Overall F1} & 
     \multicolumn{14}{c}{A:98.19; B:96.23} &
     \multicolumn{14}{c}{A:96.47; B:95.33} &
     \multicolumn{14}{c}{A:97.68; B:96.77}  &
     \multicolumn{14}{c}{A:96.23; B:94.83} &
     \multicolumn{14}{c}{A:97.09; B:96.44}&
     \multicolumn{14}{c}{A:95.77; B:93.21}&
     \multicolumn{14}{c}{A:97.49; B:96.47}\\
    \cmidrule(r){1-1}\cmidrule(lr){2-15}\cmidrule(lr){16-29}\cmidrule(lr){30-43}\cmidrule(lr){44-57}\cmidrule(lr){58-71}\cmidrule(lr){72-85}\cmidrule(lr){86-99}
    
  & \multicolumn{14}{c}{\multirow{2}[2]{*}{\includegraphics[scale=0.25]{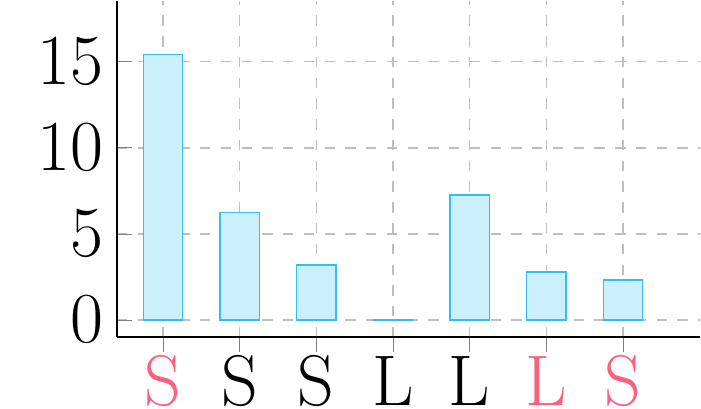}}}              & \multicolumn{14}{c}{\multirow{2}[2]{*}{\includegraphics[scale=0.25]{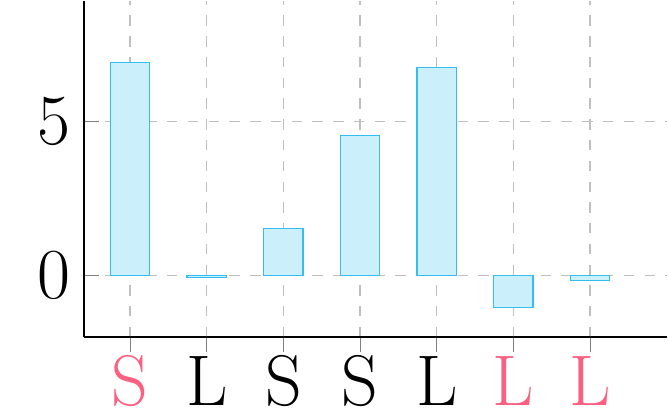}}}              & \multicolumn{14}{c}{\multirow{2}[2]{*}{\includegraphics[scale=0.25]{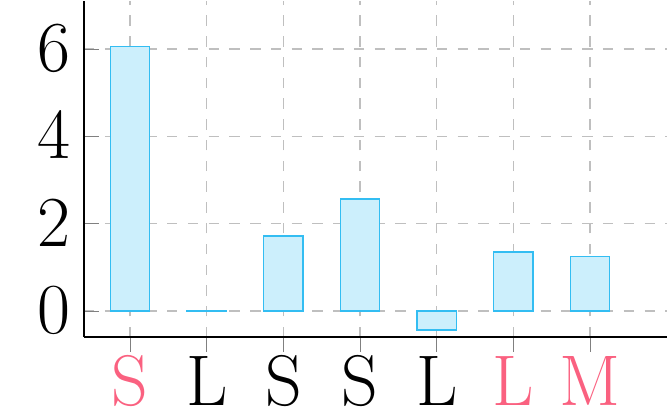}}}              & \multicolumn{14}{c}{\multirow{2}[2]{*}{\includegraphics[scale=0.25]{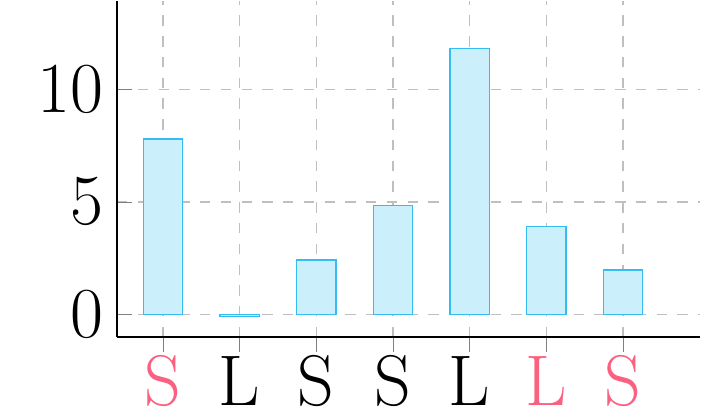}}} &
    \multicolumn{14}{c}{\multirow{2}[2]{*}{\includegraphics[scale=0.25]{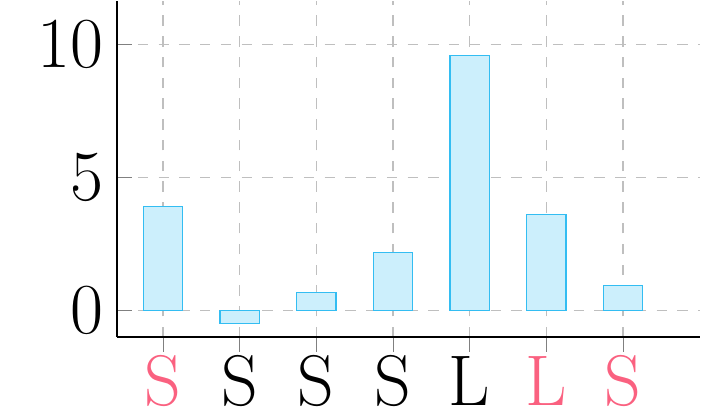}}} &
    \multicolumn{14}{c}{\multirow{2}[2]{*}{\includegraphics[scale=0.25]{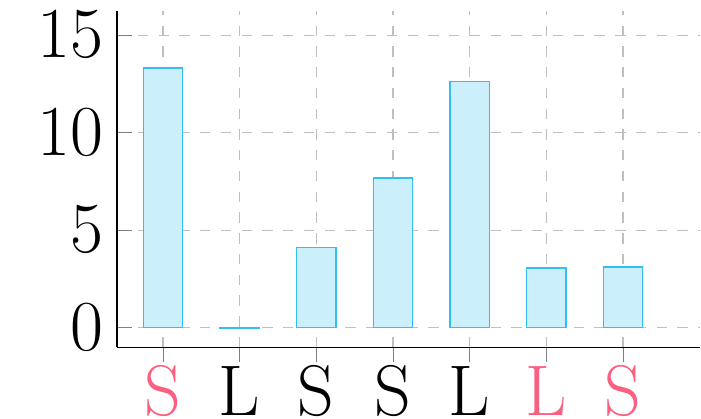}}} &
    \multicolumn{14}{c}{\multirow{2}[2]{*}{\includegraphics[scale=0.25]{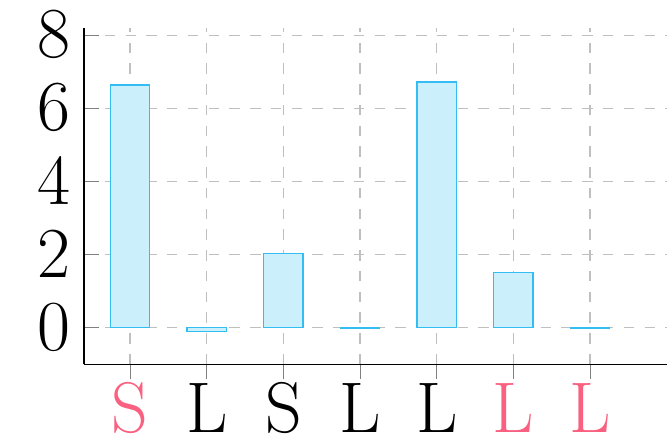}}} 
    \\
     \multicolumn{1}{l}{A: \textit{C\textcolor{myblue2}{bert}BnonLstmMlp}} &   
    \\
    \multicolumn{1}{l}{B: \textit{C\textcolor{myblue2}{elm}BnonLstmMlp}} & 
    \\  \\
    \multicolumn{1}{c}{\textcolor{brinkpink}{Aided-diagnosis}} & 
    \\ \\

    \bottomrule
    \end{tabular}%
    \vspace{-6pt}
  \caption{Diagnosis of different CWS systems. 
  For ease of presentation, we attribute values are classified into three categories: \textbf{small}(S), \textbf{middle}(M),  and \textbf{large}(L). 
  Regarding \textit{Self-diagnosis}, the x-ticklabel represents the bucket value of a specific attribute (e.g. \texttt{wLen}: word length) on which the system has achieved the worst performance. The blue bins represent the worst performance, while red bins denote the gap between worst and best performance.
  Regarding \textit{Aided-diagnosis}, the bins below the line ``$y=0$'' represent the largest gap that model $A$ is less than model $B$. By contrast, the bins above the line ``$y=0$'' denote the largest gap that model $A$ is better than model $B$.
  x-ticklabels in red indicate that the corresponding bins will be used for analysis in Sec.~\ref{sec:model-diagmosis}.
  }
  \label{tab:bucket-wise}%
\end{table*}%

\subsection{Application: Model Diagnosis}
\label{sec:model-diagmosis}

Model diagnosis is the process of identifying where the model works well and where it worse \cite{vartak2018mistique}. 
We present two types of diagnostic methods: \textbf{\textit{self-diagnosis}} and \textbf{\textit{aided-diagnosis}}.
\textit{self-diagnosis} aims to locate the bucket on which the input model has obtained the worst performance with respect to a given attribute.
For \textit{aided-diagnosis}, supposing that the holistic performance of two models satisfies: $A > B$. Then \textit{Aided-diagnosis}(A,B) will first look for a bucket, on which the performance satisfies: $A < B$. If there is no qualified bucket, then the bucket, on which model $A$ has achieved the best performance, will be returned.

Below, we will give a diagnostic analysis of some typical models shown in  Tab.~\ref{tab:bucket-wise}. The others are shown in the Appendix. 

\paragraph{{Self-diagnosis}: BERT-based models are not impeccable.}
The first row in Tab.~\ref{tab:bucket-wise} shows the diagnosis of 
model \textit{CbertBnonLstmMlp}, in which the x-ticklabel represents the bucket value of a specific attribute (e.g. \texttt{wLen}: word length) on which system has achieved worst performance.
The blue bins represent the worst performance, while red bins denote the gap between worst and best performance.
For example, the first histogram in the first row denotes that \textit{CbertBnonLstmMlp} achieved the worst performance on attribute \texttt{wCon} with value \texttt{S}.

We observe that there is a huge performance drop on all the datasets when the test samples are with the attribute values:
\texttt{wCon=S} (low label consistency of words), \texttt{cCon=S} (low label consistency of characters), \texttt{wLen=L} (long words).
This suggests that contextualized information brought from BERT is not insufficient to deal with low label consistency and long words.
To address this challenge, more efforts should be made on learning algorithms or data augmentation strategies.

\paragraph{Aided diagnosis: BERT v.s ELMo} 
The second row in Tab.~\ref{tab:bucket-wise} shows the comparing between BERT and ELMo and we observed
1)
BERT outperforms ELMo in the bucket of \texttt{wCon=S} (low label consistency of words) a lot on all datasets, suggesting that the benefit of BERT mainly comes from the processing of low label consistency of words.
2) When the OOV density of a sentence is high enough, BERT will lose its superiority.
As shown in Tab.~\ref{tab:bucket-wise}, BERT performs worse than ELMo in the bucket of \texttt{oDen=L} on the \texttt{pku} dataset whose average OOV density ($\alpha_{oDen}^{\mu}$) is the highest one (as shown in Fig.~\ref{fig:spider} (a)). 
To explain this, we take a closer look at the testing samples in the \texttt{pku} with high OOV density: ``\texttt{仰泳100米和400米}'' (backstroke 100m and 400m), ``\texttt{10月1日，北京 } (October 1, Beijing)'' . BERT, as multi-layer Transformers, is challenging to collect sufficient context to understand these cases.
3) BERT is inferior to ELMo in dealing with long sentences.
As shown in Tab.~\ref{tab:bucket-wise}, BERT obtain lower performance in the bucket of \texttt{sLen=L} on \texttt{pku} and \texttt{sxu} datasets, whose average lengths  ($\alpha_{sLen}^{\mu}$) are the highest two. 


\section{Investigation on Cross-dataset Setting}

The above in-dataset analysis aims to interpret model bias and dataset bias based on individual datasets.
In many real-world scenarios, we need to transfer a trained model to a new dataset or domain, which requires us to understand the cross-dataset generalization behavior of current systems.
In this section, our investigation on cross-dataset generalization is driven by two questions:
1) How different architectures (i.e. \textit{Cw2vBavgLstmCrf}) of CWS systems influence their cross-dataset generalization ability?
2) 
Now that we have found the common factor (\texttt{label consistency}) that affects model performance across different datasets in the previous section, can we design a measure based on it and use it to interpret cross-data generalization?
We will detail our exploration below.

\subsection{Setup}
This section focuses on the \textit{zero-shot} setting: a model with specified architecture trained on one dataset (e.g. \texttt{pku})  will be evaluated on a range of other datasets (e.g. \texttt{ctb}).
To better understand the generalization behavior of CWS systems and the relation between different datasets, we first define several measures to quantify our observations.

\subsection{Measures}
Similar to Sec.\ref{sec:indomain-measure}, 
we refer to $N_d$ as the number of all datasets and $N_m$ as the number of architectures. The cross-dataset performance can be recorded by the following matrix:
\begin{equation}
   \mathbf{U} \in \mathbb{R}^{N_d\times N_d \times N_m }
   \label{eq:generali}
\end{equation}
\vspace{-18pt}

\paragraph{Quantifying System's Cross-dataset Generalization}
Intuitively, $\mathbf{U}_{ijk} = 0.65$ represents that we have adopted the architecture $k$ (i.e. \textit{Cw2vBavgLstmCrf}) to learn a model on the training set of $i$ (e.g. \texttt{pku}), and the performance on test set of $j$ (e.g., \texttt{msr}) is $0.65$.

We do some simple numerical processing on matrix $\mathbf{U}$ to make the meaning of variables more intuitive: $           \hat{\mathbf{U}}_{ijk} = (\mathbf{U}_{jjk} - \mathbf{U}_{ijk})/\mathbf{U}_{jjk}$.
 $\hat{\mathbf{U}}_{pku,msr,k} = 0.2$ suggests that, both tested on \texttt{msr}, the model with architecture $k$ trained on \texttt{pku} is relatively lower than that trained on \texttt{msr}
by 0.2.
Usually, a lower value of $\hat{\mathbf{U}}$ is suggestive of better zero-shot generalization ability.



\paragraph{Quantifying Discrepancies of Cross-dataset Criterion}
To measure the discrepancy of segmentation criteria between any pair of training data $D^{tr}_{A}$ and test data $D^{te}_{B}$, we extend the \textit{label consistency of word} (defined in Sec.~\ref{sec:fine-grained}) to corpus-level by computing its expectation on a given training-test dataset pair. Base on Eq.~\ref{eq:def-rho}, we defined the measure $\Psi$ as:

{\small
\begin{align}
    \Psi(D^{tr}_{A}, D^{te}_{B}) = \sum_{i \in N_w}\psi(w_i^{te,k}, D^{tr}_{A}) * \mathrm{freq}(w_i^{te,k})
    \label{equ:psi}
\end{align}
}
in which $\psi(\cdot)$ (defined in Eq.~\ref{eq:def-rho}) is a function to calculate the label consistency for a test word $w_i^{te,k}$. 
$N_w$ denotes the number of unique test words and $\mathrm{freq}(w_i^{te,k})$ is the frequency of the test word. 

A lower value of $\Psi(D^{tr}_{A}, D^{te}_{B})$ suggests a larger discrepancy between the two datasets. For example,  $\Psi(D^{tr}_{msr}, D^{te}_{msr})= 78.0$ and $\Psi(D^{tr}_{msr}, D^{te}_{pku})= 75.5$, indicating that the discrepancy between \texttt{msr}'s training set and \texttt{msr}'s test set is smaller than  the discrepancy between \texttt{msr}'s training set and \texttt{pku}'s test set.  

\renewcommand\arraystretch{1.2}
\renewcommand\tabcolsep{0.9pt}
\begin{table}[!htb]
  \centering \scriptsize 
    \begin{tabular}{lcccccccccccccccc}
    \toprule
    \multicolumn{1}{c}{\multirow{2}[6]{*}{\textbf{Data}}} &  \multicolumn{8}{c}{\textbf{Data-wise ($\Psi$)}}            & \multicolumn{8}{c}{\textbf{Model-wise ($\hat{U}$)}}    \\
\cmidrule(lr){2-9} \cmidrule(lr){10-17}          & \multicolumn{1}{c}{\rotatebox{90}{\textbf{msr} } }
& \multicolumn{1}{c}{\rotatebox{90}{\textbf{pku} } } & \multicolumn{1}{c}{\rotatebox{90}{\textbf{ctb} }}  & \multicolumn{1}{c}{\rotatebox{90}{\textbf{ckip} } } & \multicolumn{1}{c}{\rotatebox{90}{\textbf{cityu} }} & \multicolumn{1}{c}{\rotatebox{90}{\textbf{ncc} }} & \multicolumn{1}{c}{\rotatebox{90}{\textbf{sxu} }}  & \multicolumn{1}{c}{\rotatebox{90}{\textcolor[rgb]{ .753,  .314,  .302}{avg}}}
& 
\multicolumn{1}{c}{\rotatebox{90}{\textbf{msr} } }
& \multicolumn{1}{c}{\rotatebox{90}{\textbf{pku} } } & \multicolumn{1}{c}{\rotatebox{90}{\textbf{ctb} }}  & \multicolumn{1}{c}{\rotatebox{90}{\textbf{ckip} } } & \multicolumn{1}{c}{\rotatebox{90}{\textbf{cityu} }} & \multicolumn{1}{c}{\rotatebox{90}{\textbf{ncc} }} & \multicolumn{1}{c}{\rotatebox{90}{\textbf{sxu} }}  & \multicolumn{1}{c}{\rotatebox{90}{\textcolor[rgb]{ .753,  .314,  .302}{avg}}}
\\
    \midrule
    \textbf{msr} & \textbf{78.0}  & 69.8  & 67.6  & 64.4  & 72.9  & 67.8  & 71.2     & \multicolumn{1}{>{\columncolor{mypink}}l}{70.2}   & 0     & 9.8   & 14    & 13    & 7.5   & 5.8   & 9.5   & \multicolumn{1}{>{\columncolor{mypink}}l}{8.4}   \\
    \textbf{pku} & 75.5  & \textbf{77.3}  & 71.9  & 66.2  & 75.6  & 68.1  & 74.5  & \multicolumn{1}{>{\columncolor{mypink}}l}{72.8}   & 11    & 0     & 7.5   & 8.3   & 3.8   & 7.6   & 5.3   & \multicolumn{1}{>{\columncolor{mypink}}l}{6.2}   \\
    \textbf{ctb} & 72.5  & 71.7  & \textbf{77.4}  & 71.0  & \textbf{76.4}  & 67.5  & 74.7  &  \multicolumn{1}{>{\columncolor{mypink}}l}{73.0}   & 14    & 8.1   & 0     & 4.1   & 2.1   & 10    & 5.8   & \multicolumn{1}{>{\columncolor{mypink}}l}{6.4}   \\
    \textbf{ckip} & 69.2  & 67.7  & 73.1  & \textbf{74.1}  & 73.3  & 67.0  & 71.2  &  \multicolumn{1}{>{\columncolor{mypink}}l}{70.8}   & 16    & 10    & 4.4   & 0     & 4.0     & 9.9   & 7.7   & \multicolumn{1}{>{\columncolor{mypink}}l}{7.4}   \\
    \textbf{cityu} & 70.2  & 67.8  & 73.3  & 70.0  & \textbf{76.3}  & 65.9  & 72.3  &  \multicolumn{1}{>{\columncolor{mypink}}l}{70.8}   & 14    & 10    & 5.2   & 5.1   & 0     & 9.2   & 6.2   & \multicolumn{1}{>{\columncolor{mypink}}l}{7.1}  \\
    \textbf{ncc} & 74.2  & 70.5  & 70.0  & 68.2  & 73.6  & \textbf{74.3}  & 73.4  &  \multicolumn{1}{>{\columncolor{mypink}}l}{72.0}   & 11    & 11    & 12    & 10    & 7.8   & 0     & 7.7   & \multicolumn{1}{>{\columncolor{mypink}}l}{8.5}   \\
    \textbf{sxu} & 72.6  & 72.1  & 71.4  & 66.9  & 75.5  & 69.1  & \textbf{78.1}  &  \multicolumn{1}{>{\columncolor{mypink}}l}{72.2}   & 13    & 7.4   & 7.5   & 8.1   & 3.0     & 8.1   & 0     & \multicolumn{1}{>{\columncolor{mypink}}l}{6.8}   \\
    \midrule
    \textcolor[rgb]{ .753,  .314,  .302}{avg} & \multicolumn{1}{>{\columncolor{mygreen}}l}{73.2}    & \multicolumn{1}{>{\columncolor{mygreen}}l}{71.0}   & \multicolumn{1}{>{\columncolor{mygreen}}l}{72.1}   & \multicolumn{1}{>{\columncolor{mygreen}}l}{68.7}   & \multicolumn{1}{>{\columncolor{mygreen}}l}{74.8}   & \multicolumn{1}{>{\columncolor{mygreen}}l}{68.5}   & \multicolumn{1}{>{\columncolor{mygreen}}l}{73.6}   & \fcolorbox{red}{white}{71.7}   & \multicolumn{1}{>{\columncolor{mygreen}}l}{11}    & \multicolumn{1}{>{\columncolor{mygreen}}l}{8.1}   & \multicolumn{1}{>{\columncolor{mygreen}}l}{7.2}   & \multicolumn{1}{>{\columncolor{mygreen}}l}{7.0}     & \multicolumn{1}{>{\columncolor{mygreen}}l}{4.0}     & \multicolumn{1}{>{\columncolor{mygreen}}l}{7.3}   & \multicolumn{1}{>{\columncolor{mygreen}}l}{6.0}     & \fcolorbox{red}{white}{7.3}  \\
    \bottomrule
    \end{tabular}%
    \vspace{-6pt}
  \caption{The relationship between different pairs of datasets measured by data-wise $\Psi$  and model-wise $\hat{U}_{k}$. Here $k$ represents the model \textit{Cw2vBavgLstmCrf}. }  \label{tab:data-bias-psi}%
\end{table}%

\subsection{Analysis}
Tab.~\ref{tab:data-bias-psi} illustrate the relationship between different train-test pair using data-wise $\Psi$  and model-wise $\hat{U}_{k}$. To test whether the expectation of label consistency is a factor that can be used to characterize cross-dataset generalization, we perform a Friedman test at $p=0.05$. Each group of samples for significance testing is obtained by changing the test-set for a given train-set (
we have 7 groups of testing samples corresponding to the 7 columns data of $\Psi$ in Tab.~\ref{tab:data-bias-psi}). The testing result is $p=0.011<0.05$, therefore, $\Psi$ can be utilized to describe the feature of a cross-dataset pair.

\paragraph{The distance between different datasets can be quantitatively characterized by $\Psi$.} 
1) As shown in Tab.~\ref{tab:data-bias-psi}, nearly all highest values are achieved on the diagonal except the row of \texttt{cityu}. $\Psi($\texttt{cityu}, \texttt{cityu}$)$ is slightly lower than $\Psi($\texttt{ctb}, \texttt{cityu}$)$, indicating the training sets of \texttt{ctb} and \texttt{cityu} are quite close.
As shown in Fig.~\ref{fig:ts-bias}(a), we do find \texttt{cityu} is closet to \texttt{ctb}.
2) \texttt{ctb} achieves the highest value in the ``avg''-column of Tab.~\ref{tab:data-bias-psi} in red, which shows taking \texttt{ctb} as the source domain, the average distance to test sets of other corpora is the smallest.
Similarly, if \texttt{cityu} is regarded as the target domain, then the average distance from other training sets to it is the smallest.
3) As shown in Fig.~\ref{fig:ts-bias}(a), \texttt{sxu}, \texttt{cityu}, and \texttt{ctb} cluster together, surrounded by other datasets \texttt{ckip}, \texttt{ncc}, and \texttt{pku} remotely, suggesting that these neighbor datasets have the similar distribution.

\paragraph{The measure $\Psi$ could be used to interpret the domain shift.}
As shown in Tab.~\ref{tab:data-bias-psi}, we find the value of ${\Psi}$ could reflect the changing trends of $\hat{U}$.
Similarly, as shown in Fig.\ref{fig:ts-bias},  impressively, these two graphs obtained in totally different ways are so close: Fig.\ref{fig:ts-bias} (a) is computed purely based on intrinsic statistics of the dataset, while Fig.\ref{fig:ts-bias} (b) is obtained based on model outputs. These qualitative results show our proposed measure $\Psi$ could be used to explain the discrepancies across datasets.

To get a more convincing observation, we additionally conduct a quantitative analysis. Specifically, we calculate the Spearman's rank correlation coefficient between $\Psi$ and the $U_k$.
The results all shown in Fig.~\ref{fig:models_results} (a-c). Encouragingly, we find that no matter which CWS system, the cross-dataset performances of them are highly correlated with our proposed measure of $\Psi$.

  {\small
\begin{figure}[!t]
    \centering 
    \subfloat[Data-wise ($\Psi$)]{
    \includegraphics[width=0.4\linewidth]{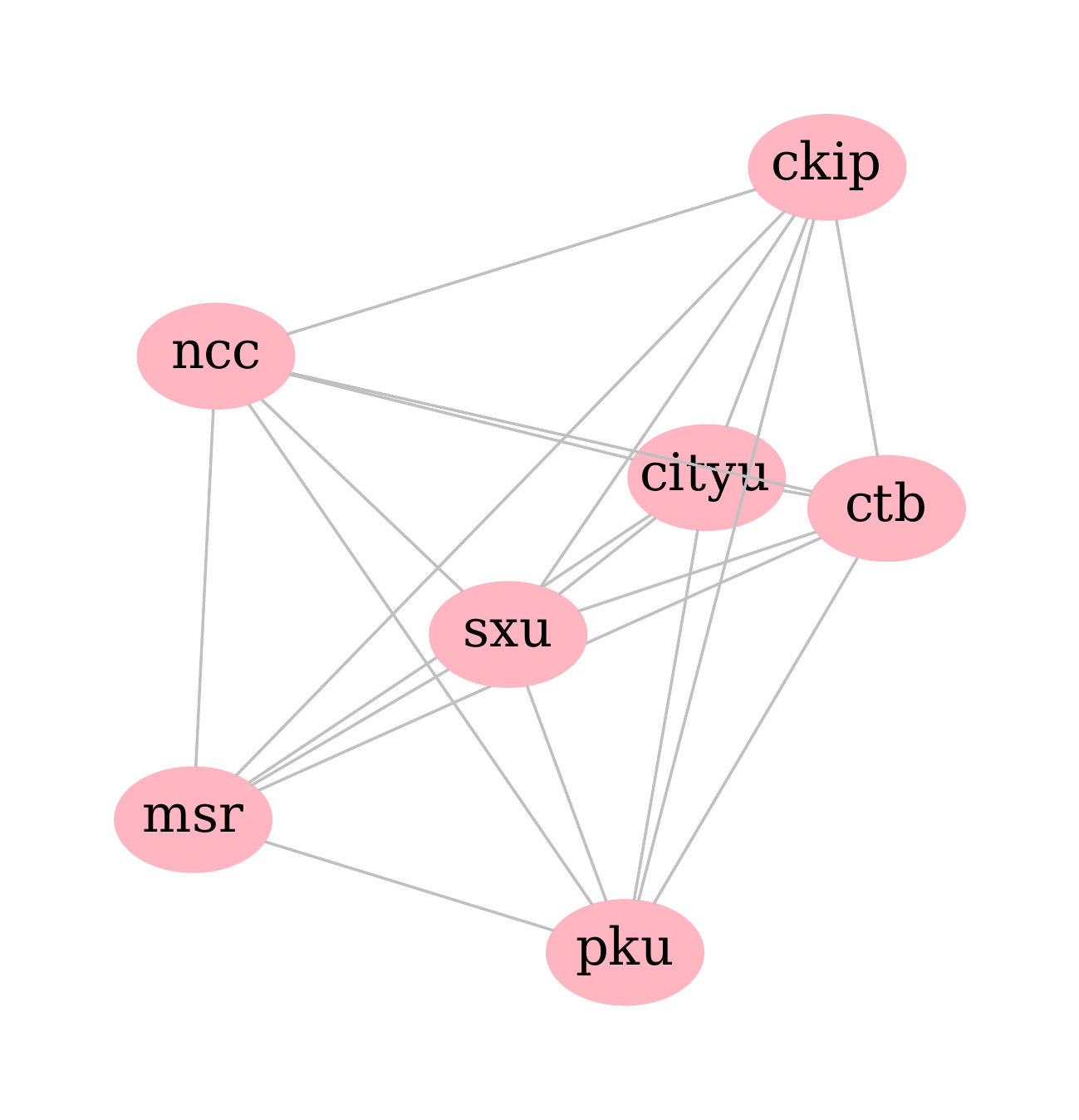} 
    }  \hspace{0.15em}  
    \subfloat[Model-wise (${U}$)]{
    \includegraphics[width=0.53\linewidth]{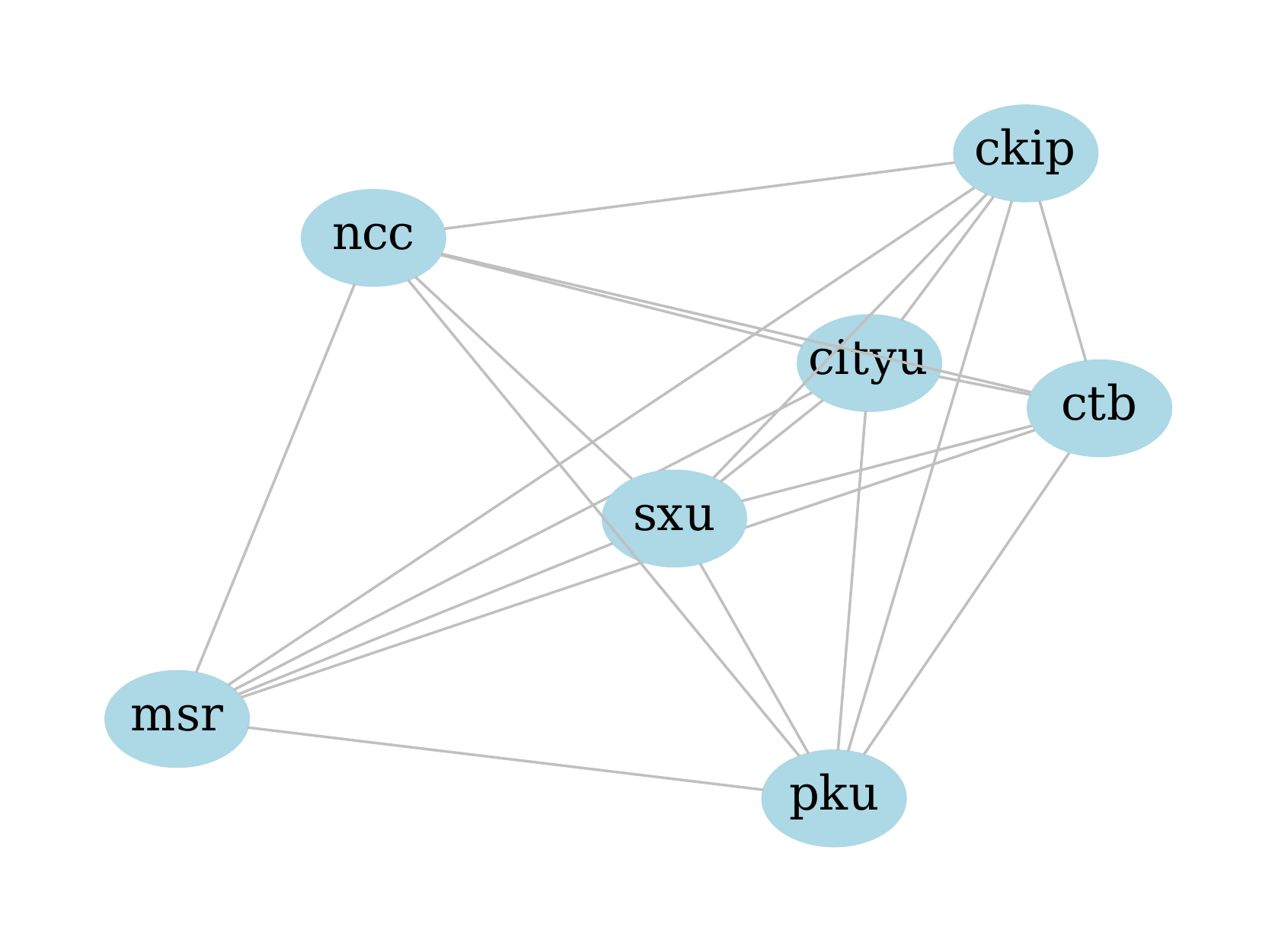}
    } 
    \vspace{-6pt}
    \caption{2D-visualization of the distances between datasets computed based on data-wise measure $\Psi$ and model-wise ${U}$ averaging on seven datasets, respectively. 
    The weight of between dataset $i$ and $j$ is transformed into an undirected edge based on:
$\frac{Z_{ij}}{Z_{jj}} +\frac{Z_{ji}}{Z_{ii}}$ and $Z$ can be $\Psi$ and $U$, in which the distance computed based on ${U}$ is the average on eight models.
    }
    \label{fig:ts-bias}
\end{figure}
}

\begin{figure}[!t]
    \centering \footnotesize
    \subfloat[Cw2vBavg]{
    \includegraphics[width=0.3\linewidth]{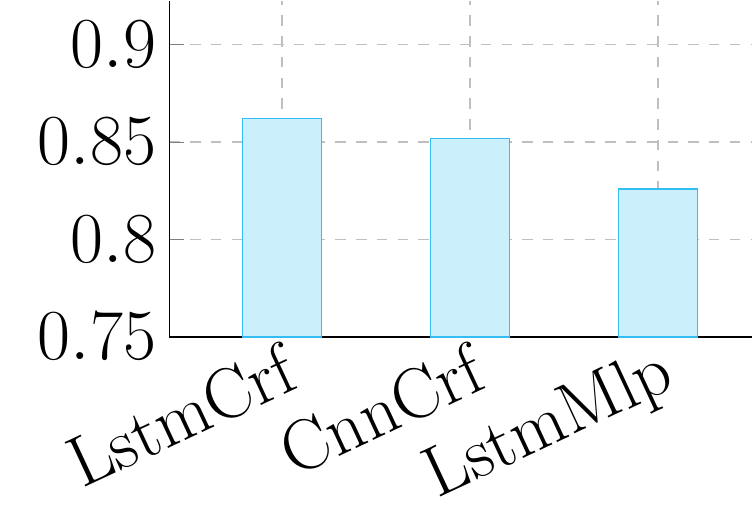} 
    }  \hspace{0.03em}  
    \subfloat[LstmCrf]{
    \includegraphics[width=0.307\linewidth]{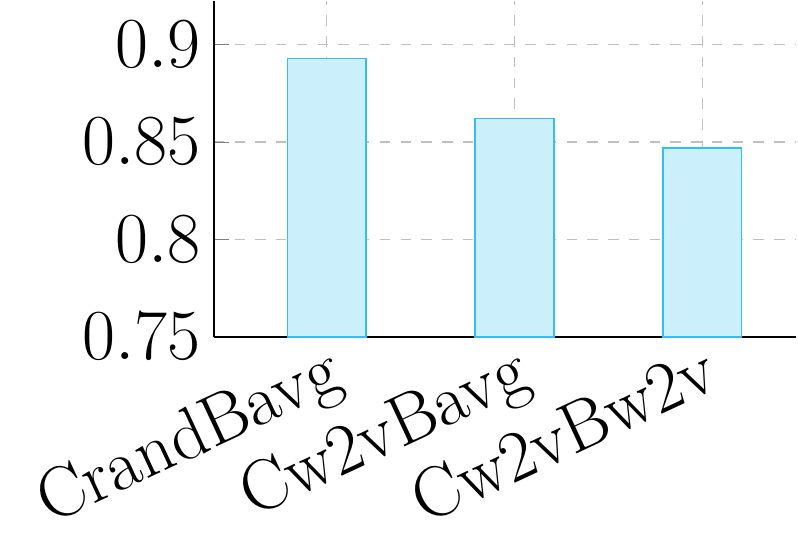} 
    }
    \hspace{0.03em}
    \subfloat[LstmMlp]{
    \includegraphics[width=0.307\linewidth]{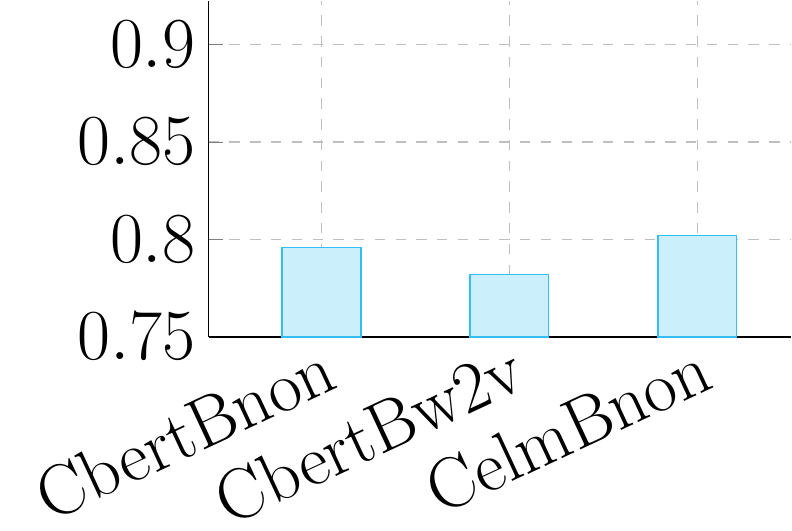}
    } 
    \vspace{-6pt}
    \caption{The Spearman's rank correlation coefficient between $\Psi$ and the $U_k$.} 
    \label{fig:models_results}
\end{figure}

\begin{algorithm}[t]
\footnotesize
\caption{Decoding Process for Dataset Order}
\label{alg:metanet}
\begin{algorithmic}[1]
{
\Require Target domain $D_t = \{D_t^{tr}, D_t^{dev}\}$; a sequence of source domains $\{D_{s_1},D_{s_2},\ldots, D_{s_N}\}$; indexes of source domains  $K=\lbrace 1 \cdots  N \rbrace;  $measure $\Phi$ 
\Require{ $\hat{K} \leftarrow \{\}$;  $\hat{D} \leftarrow D_t^{tr}$} 
\For{$k \in K$}
        \If{\texttt{Max-select}}
        \State{  $ \hat{k} = \mathrm{argmax}_{k \in K \bigwedge k\not\in \hat{K}}\Phi(\hat{D}+D_{s_k},
        D_{t}^{dev})$}
    \ElsIf{\texttt{Min-select}}
        \State{  $ \hat{k} = \mathrm{argmin}_{k \in K \bigwedge k\not\in \hat{K}}\Phi(\hat{D}+D_{s_k},
        D_{t}^{dev})$}
    \ElsIf{\texttt{Rand-select}}
        \State{  $ \hat{k} = \mathrm{Random}_{k \in K \bigwedge k\not\in \hat{K}}\Phi(\hat{D}+D_{s_k},
        D_{t}^{dev})$}
    \EndIf

    \State{$\hat{K} = \hat{K} + \{\hat{k}\}$}     \Comment{\textcolor[rgb]{0.00,0.59,0.00}{EnQueue}}
    \State{$\hat{D} = \hat{D} + D_{s_{\hat{k}}}$}
\EndFor

\Return $\hat{K}$

}
  \end{algorithmic} 
  \label{alg:select}
\end{algorithm}

\subsection{Application: Multi-source Transfer}
\label{sec:order-matter}
Given a target domain $D_t$, the above quantitative and qualitative analysis shows that the measure $\Psi$ can be used to quantify the importance of different source domains $D_{s_1},\cdots,D_{s_N}$, therefore allowing us to select suitable ones for data augmentation.

Next, we will show how to use the $\Psi$ to make better choices of source domains from the other candidates.
We take \texttt{ctb} as the tested object and continuously increase the training samples of above the seven datasets in three different ways: \textit{Rand}-, \textit{Max}-, and \textit{Min-select}.
Alg.~\ref{alg:select} shows the decoding process for the dataset order.
We choose the multi-criteria  segmenter proposed by \citet{chen2017adversarial} as our training framework for multiple datasets.

\begin{figure}[t]
    \centering
    \includegraphics[width=0.30\textwidth]{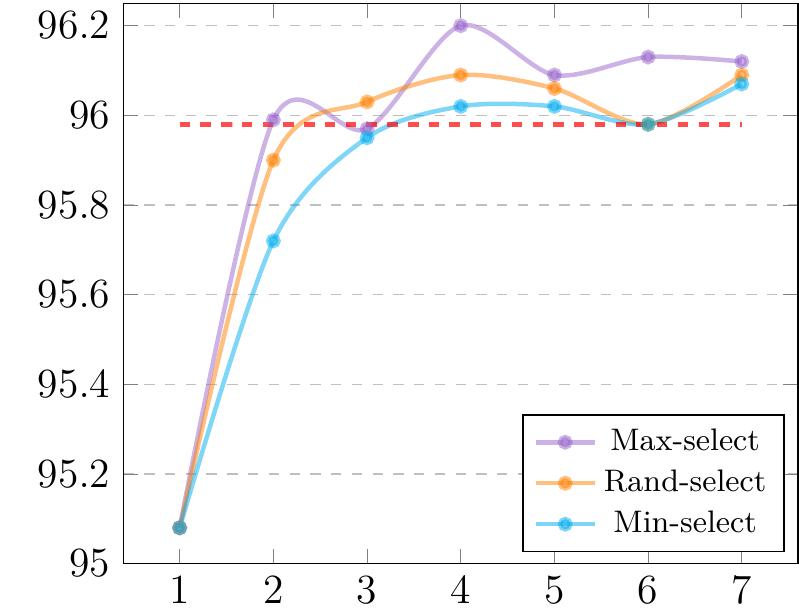}
    \vspace{-6pt}
    \caption{The changing of F1-score as more source domains are introduced in three different orders: \textit{Max-},  \textit{Min-}, and \textit{Rand-select}. The red dotted line is the result reported by \citet{chen2017adversarial} with the same model, trained on nine datasets.\protect\footnotemark[1]
    }
    \label{fig:order}
\end{figure}

\paragraph{Result}
Fig.~\ref{fig:order} illustrates the changes in F1-score as more source domains are introduced in three different orders. We do a Friedman test with the null hypothesis that the order of training set introduced had no influence on the performance of a given model. The significance testing result shows that the training set introduced with Max-, Min-, and Rand-select are significantly different ($p=8.0\times10^{-3}<0.05$).  
We can observe from Fig.~\ref{fig:order} that:
\textit{More training samples are not a guarantee of better results for CWS models due to the criteria discrepancy between different datasets.}

Specifically, the \textit{Max-select} operation helps us find an optimal set of source domains (\texttt{ctb}, \texttt{sxu}, \texttt{ncc}, \texttt{cityu}), on which the model could achieve the best results, outperforming \citet{chen2017adversarial}'s result by a significant margin, which trained on nine datasets (two more than ours).  
Regarding the two baseline decoding strategies (\textit{Min-select} and \textit{Rand-select}), we find the best performance on \texttt{ctb} are both obtained when all seven training sets are used.
The above observations indicate that, when we introduce multiple training sets for data augmentation, the order of the distance between training and development sets can help us select which parts of source domains are useful.
And $\Psi$, we proposed in this paper,  is an effective measure to quantify this order (without learning process), providing a novel solution for multi-source transfer learning.

\footnotetext[1]{To make a fair comparison, all results are implemented based on their public code.}

\section{Discussion}


We summarize the main observations from our experiments and try to give preliminary answers to our proposed research questions:

\noindent\textbf{\emph{Does existing excellent performance imply a perfect CWS system?}} No. Beyond giving this unsurprising answer, we present an interpretable evaluation method to help us diagnose the weaknesses of existing top-performing systems and relative merits between two systems. For example, we find even top-scoring BERT-based models still cannot deal with the words with low label consistency or long words well, and BERT is inferior to ELMo as an encoder in dealing with long sentences.

\noindent\textbf{\emph{Is  there  a  one-size-fits-all  system?}} No (Best-performing systems on different datasets frequently involve diverse neural architectures).
Although this question can be answered relatively easily by simply looking at the overall results of different systems in diverse data sets (Sec.\ref{tab:holistic}), we take a step further to how to make choices of them (\texttt{BERT v.s ELMo},  \texttt{LSTM v.s CNN})) by conducting dataset bias-aware \textit{Aided-diagnosis} (Sec.\ref{sec:model-diagmosis}). 

\noindent\textbf{\emph{Can we design a measure to quantify the discrepancies among different criteria?}}
Yes. 
We first verify that the \textit{label consistency} of words and \textit{word length} have a more consistent impact on CWS performance. Based on this, we design a measure to quantify the distance between different datasets, which correlates well with the cross-dataset performance and can be used for multi-source transfer learning, help us avoid the negative transfer.

\section*{Acknowledgments}
The authors wish to thank Zhenghua Li and Meishan Zhang for their helpful comments and carefully proofreading of our manuscript. This work was partially funded by China National Key R\&D Program (No. 2017YFB1002104，2018YFC0831105, 2018YFB1005104), National Natural Science Foundation of China (No. 61976056, 61751201, 61532011), Shanghai Municipal Science and Technology Major Project (No.2018SHZDZX01), Science and Technology Commission of Shanghai Municipality Grant  (No.18DZ1201000, 17JC1420200).

\end{CJK*}

\bibliography{nlp}

\begin{thebibliography}{31}
\expandafter\ifx\csname natexlab\endcsname\relax\def\natexlab#1{#1}\fi

\bibitem[{Cai and Zhao(2016)}]{cai2016neural}
Deng Cai and Hai Zhao. 2016.
\newblock Neural word segmentation learning for chinese.
\newblock In \emph{Proceedings of the 54th Annual Meeting of the Association
  for Computational Linguistics (Volume 1: Long Papers)}, pages 409--420.

\bibitem[{Chen et~al.(2015{\natexlab{a}})Chen, Qiu, Zhu, and
  Huang}]{chen2015gated}
Xinchi Chen, Xipeng Qiu, Chenxi Zhu, and Xuanjing Huang. 2015{\natexlab{a}}.
\newblock Gated recursive neural network for chinese word segmentation.
\newblock In \emph{Proceedings of the 53rd Annual Meeting of the Association
  for Computational Linguistics and the 7th International Joint Conference on
  Natural Language Processing (Volume 1: Long Papers)}, pages 1744--1753.

\bibitem[{Chen et~al.(2015{\natexlab{b}})Chen, Qiu, Zhu, Liu, and
  Huang}]{chen2015long}
Xinchi Chen, Xipeng Qiu, Chenxi Zhu, Pengfei Liu, and Xuanjing Huang.
  2015{\natexlab{b}}.
\newblock Long short-term memory neural networks for chinese word segmentation.
\newblock In \emph{Proceedings of the 2015 Conference on Empirical Methods in
  Natural Language Processing}, pages 1197--1206.

\bibitem[{Chen et~al.(2017)Chen, Shi, Qiu, and Huang}]{chen2017adversarial}
Xinchi Chen, Zhan Shi, Xipeng Qiu, and Xuanjing Huang. 2017.
\newblock Adversarial multi-criteria learning for chinese word segmentation.
\newblock In \emph{Proceedings of the 55th Annual Meeting of the Association
  for Computational Linguistics (Volume 1: Long Papers)}, pages 1193--1203.

\bibitem[{Collobert et~al.(2011)Collobert, Weston, Bottou, Karlen, Kavukcuoglu,
  and Kuksa}]{collobert2011natural}
Ronan Collobert, Jason Weston, L{\'e}on Bottou, Michael Karlen, Koray
  Kavukcuoglu, and Pavel Kuksa. 2011.
\newblock Natural language processing (almost) from scratch.
\newblock \emph{The Journal of Machine Learning Research}, 12:2493--2537.

\bibitem[{Devlin et~al.(2018)Devlin, Chang, Lee, and
  Toutanova}]{devlin2018bert}
Jacob Devlin, Ming-Wei Chang, Kenton Lee, and Kristina Toutanova. 2018.
\newblock Bert: Pre-training of deep bidirectional transformers for language
  understanding.
\newblock \emph{arXiv preprint arXiv:1810.04805}.

\bibitem[{Fu et~al.(2020{\natexlab{a}})Fu, Liu, and
  Neubig}]{fu2020interpretable}
Jinlan Fu, Pengfei Liu, and Graham Neubig. 2020{\natexlab{a}}.
\newblock Interpretable multi-dataset evaluation for named entity recognition.
\newblock \emph{arXiv preprint arXiv:2011.06854}.

\bibitem[{Fu et~al.(2020{\natexlab{b}})Fu, Liu, Zhang, and
  Huang}]{fu2020rethinking}
Jinlan Fu, Pengfei Liu, Qi~Zhang, and Xuanjing Huang. 2020{\natexlab{b}}.
\newblock Rethinking generalization of neural models: A named entity
  recognition case study.
\newblock In \emph{AAAI}, pages 7732--7739.

\bibitem[{Gong et~al.(2017)Gong, Li, Zhang, and Jiang}]{gong2017multi}
Chen Gong, Zhenghua Li, Min Zhang, and Xinzhou Jiang. 2017.
\newblock Multi-grained chinese word segmentation.
\newblock In \emph{Proceedings of the 2017 Conference on Empirical Methods in
  Natural Language Processing}, pages 692--703.

\bibitem[{Hochreiter and Schmidhuber(1997)}]{hochreiter1997long}
Sepp Hochreiter and J{\"u}rgen Schmidhuber. 1997.
\newblock Long short-term memory.
\newblock \emph{Neural computation}, 9(8):1735--1780.

\bibitem[{Huang et~al.(2007)Huang, {\v{S}}imon, Hsieh, and
  Pr{\'e}vot}]{huang2007rethinking}
Chu-Ren Huang, Petr {\v{S}}imon, Shu-Kai Hsieh, and Laurent Pr{\'e}vot. 2007.
\newblock Rethinking chinese word segmentation: tokenization, character
  classification, or wordbreak identification.
\newblock In \emph{Proceedings of the 45th annual meeting of the ACL on
  interactive poster and demonstration sessions}, pages 69--72. Association for
  Computational Linguistics.

\bibitem[{Huang et~al.(2019)Huang, Cheng, Chen, Wang, and
  Chu}]{huang2019toward}
Weipeng Huang, Xingyi Cheng, Kunlong Chen, Taifeng Wang, and Wei Chu. 2019.
\newblock Toward fast and accurate neural chinese word segmentation with
  multi-criteria learning.
\newblock \emph{arXiv preprint arXiv:1903.04190}.

\bibitem[{Kalchbrenner et~al.(2014)Kalchbrenner, Grefenstette, and
  Blunsom}]{kalchbrenner2014convolutional}
Nal Kalchbrenner, Edward Grefenstette, and Phil Blunsom. 2014.
\newblock A convolutional neural network for modelling sentences.
\newblock In \emph{Proceedings of ACL}.

\bibitem[{Lample et~al.(2016)Lample, Ballesteros, Subramanian, Kawakami, and
  Dyer}]{lample2016neural}
Guillaume Lample, Miguel Ballesteros, Sandeep Subramanian, Kazuya Kawakami, and
  Chris Dyer. 2016.
\newblock Neural architectures for named entity recognition.
\newblock In \emph{Proceedings of NAACL-HLT}, pages 260--270.

\bibitem[{Liu et~al.(2014)Liu, Zhang, Che, Liu, and Wu}]{liu2014domain}
Yijia Liu, Yue Zhang, Wanxiang Che, Ting Liu, and Fan Wu. 2014.
\newblock Domain adaptation for crf-based chinese word segmentation using free
  annotations.
\newblock In \emph{Proceedings of the 2014 Conference on Empirical Methods in
  Natural Language Processing (EMNLP)}, pages 864--874.

\bibitem[{Luo and Yang(2016)}]{luo2016empirical}
Wencan Luo and Fan Yang. 2016.
\newblock An empirical study of automatic chinese word segmentation for spoken
  language understanding and named entity recognition.
\newblock In \emph{Proceedings of the 2016 Conference of the North American
  Chapter of the Association for Computational Linguistics: Human Language
  Technologies}, pages 238--248.

\bibitem[{Ma et~al.(2018)Ma, Ganchev, and Weiss}]{ma2018state}
Ji~Ma, Kuzman Ganchev, and David Weiss. 2018.
\newblock State-of-the-art chinese word segmentation with bi-lstms.
\newblock In \emph{Proceedings of the 2018 Conference on Empirical Methods in
  Natural Language Processing}, pages 4902--4908.

\bibitem[{Mikolov et~al.(2013)Mikolov, Chen, Corrado, and
  Dean}]{mikolov2013efficient}
Tomas Mikolov, Kai Chen, Greg Corrado, and Jeffrey Dean. 2013.
\newblock Efficient estimation of word representations in vector space.
\newblock \emph{arXiv preprint arXiv:1301.3781}.

\bibitem[{Mukaka(2012)}]{mukaka2012guide}
Mavuto~M Mukaka. 2012.
\newblock A guide to appropriate use of correlation coefficient in medical
  research.
\newblock \emph{Malawi Medical Journal}, 24(3):69--71.

\bibitem[{Pei et~al.(2014)Pei, Ge, and Chang}]{pei2014max}
Wenzhe Pei, Tao Ge, and Baobao Chang. 2014.
\newblock Max-margin tensor neural network for chinese word segmentation.
\newblock In \emph{Proceedings of the 52nd Annual Meeting of the Association
  for Computational Linguistics (Volume 1: Long Papers)}, pages 293--303.

\bibitem[{Peters et~al.(2018)Peters, Neumann, Iyyer, Gardner, Clark, Lee, and
  Zettlemoyer}]{peters2018deep}
Matthew Peters, Mark Neumann, Mohit Iyyer, Matt Gardner, Christopher Clark,
  Kenton Lee, and Luke Zettlemoyer. 2018.
\newblock Deep contextualized word representations.
\newblock In \emph{Proceedings of the 2018 Conference of the North American
  Chapter of the Association for Computational Linguistics: Human Language
  Technologies, Volume 1 (Long Papers)}, volume~1, pages 2227--2237.

\bibitem[{Qian et~al.(2016)Qian, Qiu, and Huang}]{qian2016a}
Peng Qian, Xipeng Qiu, and Xuanjing Huang. 2016.
\newblock A new psychometric-inspired evaluation metric for chinese word
  segmentation.
\newblock 1:2185--2194.

\bibitem[{Qiu et~al.(2019)Qiu, Pei, Yan, and Huang}]{qiu2019multi-criteria}
Xipeng Qiu, Hengzhi Pei, Hang Yan, and Xuanjing Huang. 2019.
\newblock Multi-criteria chinese word segmentation with transformer.
\newblock \emph{arXiv: Computation and Language}.

\bibitem[{Sproat and Shih(1990)}]{sproat1990statistical}
Richard Sproat and Chilin Shih. 1990.
\newblock A statistical method for finding word boundaries in chinese text.
\newblock \emph{Computer Processing of Chinese and Oriental Languages},
  4(4):336--351.

\bibitem[{Vartak et~al.(2018)Vartak, F~da Trindade, Madden, and
  Zaharia}]{vartak2018mistique}
Manasi Vartak, Joana~M F~da Trindade, Samuel Madden, and Matei Zaharia. 2018.
\newblock Mistique: A system to store and query model intermediates for model
  diagnosis.
\newblock In \emph{Proceedings of the 2018 International Conference on
  Management of Data}, pages 1285--1300. ACM.

\bibitem[{Xue and Shen(2003)}]{xue2003chinese}
Nianwen Xue and Libin Shen. 2003.
\newblock Chinese word segmentation as lmr tagging.
\newblock In \emph{Proceedings of the second SIGHAN workshop on Chinese
  language processing-Volume 17}, pages 176--179. Association for Computational
  Linguistics.

\bibitem[{Yang et~al.(2017)Yang, Zhang, and Dong}]{yang2017neural}
Jie Yang, Yue Zhang, and Fei Dong. 2017.
\newblock Neural word segmentation with rich pretraining.
\newblock In \emph{Proceedings of the 55th Annual Meeting of the Association
  for Computational Linguistics (Volume 1: Long Papers)}, pages 839--849.

\bibitem[{Yang et~al.(2019)Yang, Zhang, and Liang}]{yang2019subword}
Jie Yang, Yue Zhang, and Shuailong Liang. 2019.
\newblock Subword encoding in lattice lstm for chinese word segmentation.
\newblock In \emph{Proceedings of the 2019 Conference of the North American
  Chapter of the Association for Computational Linguistics: Human Language
  Technologies, Volume 1 (Long and Short Papers)}, pages 2720--2725.

\bibitem[{Zheng et~al.(2013)Zheng, Chen, and Xu}]{zheng2013deep}
Xiaoqing Zheng, Hanyang Chen, and Tianyu Xu. 2013.
\newblock Deep learning for chinese word segmentation and pos tagging.
\newblock In \emph{Proceedings of the 2013 Conference on Empirical Methods in
  Natural Language Processing}, pages 647--657.

\bibitem[{Zhou et~al.(2017)Zhou, Yu, Zhang, Huang, Dai, and
  Chen}]{zhou2017word}
Hao Zhou, Zhenting Yu, Yue Zhang, Shujian Huang, Xin-Yu Dai, and Jiajun Chen.
  2017.
\newblock Word-context character embeddings for chinese word segmentation.
\newblock In \emph{Proceedings of the 2017 Conference on Empirical Methods in
  Natural Language Processing}, pages 760--766.

\bibitem[{Zimmerman and Zumbo(1993)}]{zimmerman1993relative}
Donald~W Zimmerman and Bruno~D Zumbo. 1993.
\newblock Relative power of the wilcoxon test, the friedman test, and
  repeated-measures anova on ranks.
\newblock \emph{The Journal of Experimental Education}, 62(1):75--86.

\end{thebibliography}
\bibliographystyle{acl_natbib}

\appendix

\begin{CJK*}{UTF8}{gbsn}
\section{Significance Testing}
To conduct the fine-grained evaluation, we divide the words (characters) of the test set into several subsets, which are named buckets in this paper. We perform Friedman significance testing at $p=0.05$ in  dataset-dimension and model-dimension, and the results are shown in Tab.~\ref{tab:signif-data} and Tab.~\ref{tab:signif-model}, respectively. For dataset-dimension (model-dimension),
the null hypothesis is that the performance of buckets concerning an attribute has the same means for a given dataset (model).




\begin{table*}[!ht]
  \centering \footnotesize
    \begin{tabular}{lrrrrrrr}
    \toprule
    \textbf{datas} & \multicolumn{1}{l}{\textbf{wCon}} & \multicolumn{1}{l}{\textbf{cCon}} & \multicolumn{1}{l}{\textbf{cFre}} & \multicolumn{1}{l}{\textbf{wFre}} & \multicolumn{1}{l}{\textbf{wLen}} & \multicolumn{1}{l}{\textbf{oDen}} & \multicolumn{1}{l}{\textbf{sLen}} \\
    \midrule
    msr   & $1.2\times10^{-11}$ & $1.0\times10^{-11}$ & $ 2.5\times10^{-11}$ & $9.8\times10^{-11}$ & $2.0\times10^{-10}$ & $5.8\times10^{-09}$ & $9.1\times10^{-09}$ \\
    pku   & $7.2\times10^{-12}$ & $1.1\times10^{-11}$ & $2.7\times10^{-11}$ & $1.5\times10^{-10}$ & $8.4\times10^{-11}$ & $1.4\times10^{-07}$ & $4.8\times10^{-08}$ \\
    ctb   & $7.2\times10^{-12}$ & $8.0\times10^{-12}$ & $3.9\times10^{-11}$ & $3.6\times10^{-10}$ & $1.5\times10^{-10}$ & $1.3\times10^{-07}$ & $4.9\times10^{-07}$ \\
    ckip  & $7.2\times10^{-12}$ & $7.3\times10^{-12}$ & $9.5\times10^{-10}$ & $1.0\times10^{-10}$ & $8.3\times10^{-09}$ & $2.9\times10^{-11}$ & $5.5\times10^{-05}$ \\
    cityu & $6.2\times10^{-12}$ & $1.0\times10^{-11}$ & $4.1\times10^{-11}$ & $9.6\times10^{-11}$ & $4.5\times10^{-10}$ & $8.1\times10^{-11}$ & $2.3\times10^{-10}$ \\
    ncc   & $7.2\times10^{-12}$ & $7.4\times10^{-12}$ & $7.8\times10^{-12}$ & $1.6\times10^{-10}$ & $2.6\times10^{-11}$ & $2.2\times10^{-10}$ & $1.7\times10^{-09}$ \\
    sxu   & $6.3\times10^{-12}$ & $9.3\times10^{-12}$ & $2.1\times10^{-11}$ & $1.3\times10^{-10}$ & $7.9\times10^{-09}$ & $2.6\times10^{-08}$ & $5.5\times10^{-08}$ \\
    \bottomrule
    \end{tabular}%
    \caption{$p$-values from the Friedman test. The null hypothesis is that the performance of different buckets with respect to an attribute has the same means for a given \textbf{dataset}.}
  \label{tab:signif-data}%
\end{table*}%

\renewcommand\tabcolsep{1.5pt}
\begin{table*}[!ht]
  \centering \footnotesize
    \begin{tabular}{cccccccc}
    \toprule
    \textbf{models} & \multicolumn{1}{l}{\textbf{wCon}} & \multicolumn{1}{l}{\textbf{cCon}} & \multicolumn{1}{l}{\textbf{cFre}} & \multicolumn{1}{l}{\textbf{wFre}} & \multicolumn{1}{l}{\textbf{wLen}} & \multicolumn{1}{l}{\textbf{oDen}} & \multicolumn{1}{l}{\textbf{sLen}} \\
    \midrule
    CrandBavgLstmCrf & $6.5\times10^{-10}$ & $5.4\times10^{-10}$ & $6.9\times10^{-7}$ & $9.5\times10^{-5}$ & $4.5\times10^{-5}$ & $1.8\times10^{-5}$ & \textcolor{brinkpink}{$2.1\times10^{-1}$} \\
    Cw2vBavgLstmCrf & $6.5\times10^{-10}$ & $6.6\times10^{-10}$ & $5.7\times10^{-7}$ & $7.8\times10^{-5}$ & $2.9\times10^{-5}$ & $2.8\times10^{-6}$ & \textcolor{brinkpink}{$2.0\times10^{-1}$} \\
    Cw2vBavgLstmMlp & $6.0\times10^{-10}$ & $7.5\times10^{-10}$ & $1.4\times10^{-7}$ & $1.6\times10^{-4}$ & $3.8\times10^{-5}$ & $3.3\times10^{-4}$ & \textcolor{brinkpink}{$4.5\times10^{-1}$} \\
    Cw2vBavgCnnCrf & $5.7\times10^{-10}$ & $5.1\times10^{-10}$ & $4.5\times10^{-7}$ & $5.9\times10^{-5}$ & $1.1\times10^{-4}$ & $1.1\times10^{-4}$ & \textcolor{brinkpink}{$8.3\times10^{-1}$} \\
    Cw2vBw2vLstmCrf & $6.5\times10^{-10}$ & $5.2\times10^{-10}$ & $3.0\times10^{-7}$ & $1.2\times10^{-4}$ & $2.3\times10^{-5}$ & $6.8\times10^{-6}$ & \textcolor{brinkpink}{$2.6\times10^{-1}$} \\
    CelmBnonLstmMlp & $6.6\times10^{-10}$ & $6.5\times10^{-10}$ & $1.0\times10^{-5}$ & $1.9\times10^{-4}$ & $4.5\times10^{-4}$ & $4.1\times10^{-4}$ & $2.1\times10^{-4}$ \\
    CbertBnonLstmMlp & $7.5\times10^{-10}$ & $1.4\times10^{-9}$ & $3.8\times10^{-5}$ & $1.3\times10^{-4}$ & $4.4\times10^{-5}$ & $1.1\times10^{-4}$ & $2.7\times10^{-2}$ \\
    CbertBw2vLstmMlp & $6.6\times10^{-10}$ & $7.8\times10^{-10}$ & $1.5\times10^{-5}$ & $1.1\times10^{-4}$ & $5.4\times10^{-5}$ & $8.0\times10^{-3}$ & \textcolor{brinkpink}{$6.5\times10^{-2}$} \\
    \bottomrule
    \end{tabular}%
     \caption{$p$-values from the Friedman test. The null hypothesis is that the performance of different buckets with respect to an attribute has the same means for a given \textbf{model}. The values in pink indicate that the value is greater than $p=0.05$. }
  \label{tab:signif-model}%
\end{table*}%

\renewcommand\tabcolsep{0.75pt}
\renewcommand\arraystretch{0.68}  
\begin{table*}[!htb]
  \centering \tiny
    \begin{tabular}{ccccccc ccccccc ccccccc ccccccc ccccccc ccccccc ccccccc cccccccccccccccccccccccccccccccccccccccccccccccccc}
    \toprule
          & \multicolumn{14}{c}{msr }                           & \multicolumn{14}{c}{pku}                            & \multicolumn{14}{c}{ctb}                            & \multicolumn{14}{c}{ckip}  &
          \multicolumn{14}{c}{cityu}  &
          \multicolumn{14}{c}{ncc}  &
          \multicolumn{14}{c}{sxu}\\
    \midrule
    \multicolumn{1}{l}{} & &&&&&&& \multicolumn{1}{l}{\rotatebox{90}{wCon}} & \multicolumn{1}{l}{\rotatebox{90}{cCon}} & \multicolumn{1}{l}{\rotatebox{90}{cFre}} & \multicolumn{1}{l}{\rotatebox{90}{wFre}} & \multicolumn{1}{l}{\rotatebox{90}{wLen}} & \multicolumn{1}{l}{\rotatebox{90}{oDen}} & \multicolumn{1}{l}{\rotatebox{90}{sLen}} &
    &&&&&&&
    \multicolumn{1}{l}{\rotatebox{90}{wCon}} & \multicolumn{1}{l}{\rotatebox{90}{cCon}} & \multicolumn{1}{l}{\rotatebox{90}{cFre}} & \multicolumn{1}{l}{\rotatebox{90}{wFre}} & \multicolumn{1}{l}{\rotatebox{90}{wLen}} & \multicolumn{1}{l}{\rotatebox{90}{oDen}} & \multicolumn{1}{l}{\rotatebox{90}{sLen}} &
    &&&&&&&
    \multicolumn{1}{l}{\rotatebox{90}{wCon}} & \multicolumn{1}{l}{\rotatebox{90}{cCon}} & \multicolumn{1}{l}{\rotatebox{90}{cFre}} & \multicolumn{1}{l}{\rotatebox{90}{wFre}} & \multicolumn{1}{l}{\rotatebox{90}{wLen}} & \multicolumn{1}{l}{\rotatebox{90}{oDen}} & \multicolumn{1}{l}{\rotatebox{90}{sLen}} &
    &&&&&&&
    \multicolumn{1}{l}{\rotatebox{90}{wCon}} & \multicolumn{1}{l}{\rotatebox{90}{cCon}} & \multicolumn{1}{l}{\rotatebox{90}{cFre}} & \multicolumn{1}{l}{\rotatebox{90}{wFre}} & \multicolumn{1}{l}{\rotatebox{90}{wLen}} & \multicolumn{1}{l}{\rotatebox{90}{oDen}} & \multicolumn{1}{l}{\rotatebox{90}{sLen}} &
    &&&&&&&
    \multicolumn{1}{l}{\rotatebox{90}{wCon}} & \multicolumn{1}{l}{\rotatebox{90}{cCon}} & \multicolumn{1}{l}{\rotatebox{90}{cFre}} & \multicolumn{1}{l}{\rotatebox{90}{wFre}} & \multicolumn{1}{l}{\rotatebox{90}{wLen}} & \multicolumn{1}{l}{\rotatebox{90}{oDen}} & \multicolumn{1}{l}{\rotatebox{90}{sLen}} &
    &&&&&&&
    \multicolumn{1}{l}{\rotatebox{90}{wCon}} & \multicolumn{1}{l}{\rotatebox{90}{cCon}} & \multicolumn{1}{l}{\rotatebox{90}{cFre}} & \multicolumn{1}{l}{\rotatebox{90}{wFre}} & \multicolumn{1}{l}{\rotatebox{90}{wLen}} & \multicolumn{1}{l}{\rotatebox{90}{oDen}} & \multicolumn{1}{l}{\rotatebox{90}{sLen}} &
    &&&&&&&
    \multicolumn{1}{l}{\rotatebox{90}{wCon}} & \multicolumn{1}{l}{\rotatebox{90}{cCon}} & \multicolumn{1}{l}{\rotatebox{90}{cFre}} & \multicolumn{1}{l}{\rotatebox{90}{wFre}} & \multicolumn{1}{l}{\rotatebox{90}{wLen}} & \multicolumn{1}{l}{\rotatebox{90}{oDen}} & \multicolumn{1}{l}{\rotatebox{90}{sLen}} 
    \\
    \midrule
     \multicolumn{1}{l}{Overall F1} & 
     \multicolumn{14}{c}{A:96.46; B:96.48} &
     \multicolumn{14}{c}{A:94.10; B:93.99} &
     \multicolumn{14}{c}{A:95.08; B:94.74} &
     \multicolumn{14}{c}{A:92.81; B:92.73}&
     \multicolumn{14}{c}{A:93.67; B:93.72}&
     \multicolumn{14}{c}{A:92.04; B:92.64}&
     \multicolumn{14}{c}{A:94.71; B:94.36} \\
    \cmidrule(r){1-1}\cmidrule(lr){2-15}\cmidrule(lr){16-29}\cmidrule(lr){30-43}\cmidrule(lr){44-57}\cmidrule(lr){58-71}\cmidrule(lr){72-85}\cmidrule(lr){86-99}
    
  & \multicolumn{14}{c}{\multirow{4}[2]{*}{\includegraphics[scale=0.25]{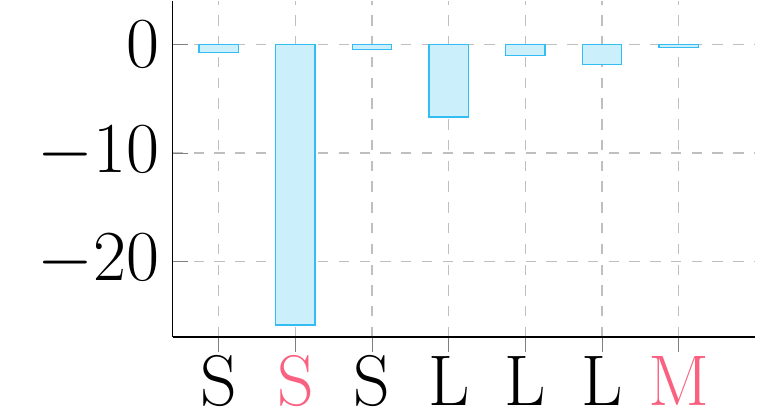}}}              & \multicolumn{14}{c}{\multirow{4}[2]{*}{\includegraphics[scale=0.25]{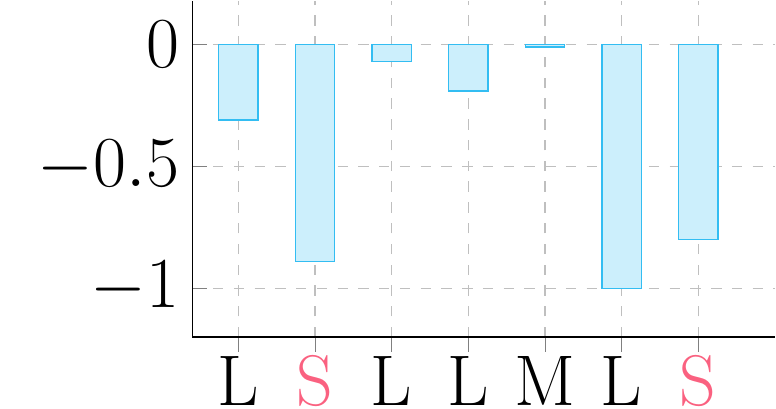}}}              & \multicolumn{14}{c}{\multirow{4}[2]{*}{\includegraphics[scale=0.25]{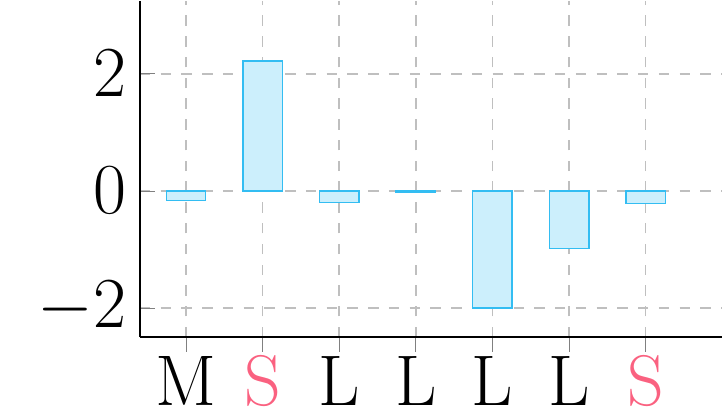}}}              & \multicolumn{14}{c}{\multirow{4}[2]{*}{\includegraphics[scale=0.25]{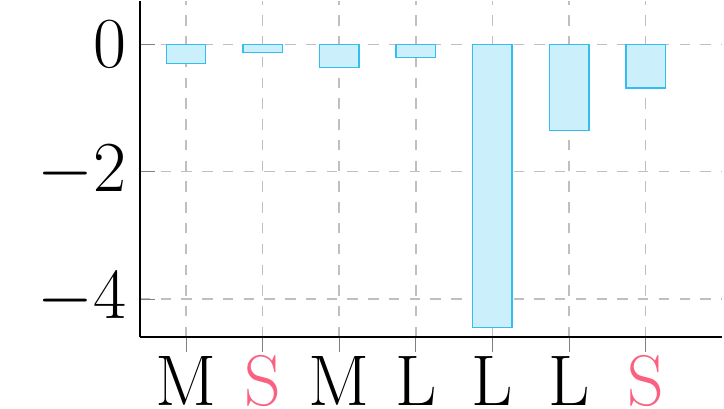}}} &
    \multicolumn{14}{c}{\multirow{4}[2]{*}{\includegraphics[scale=0.25]{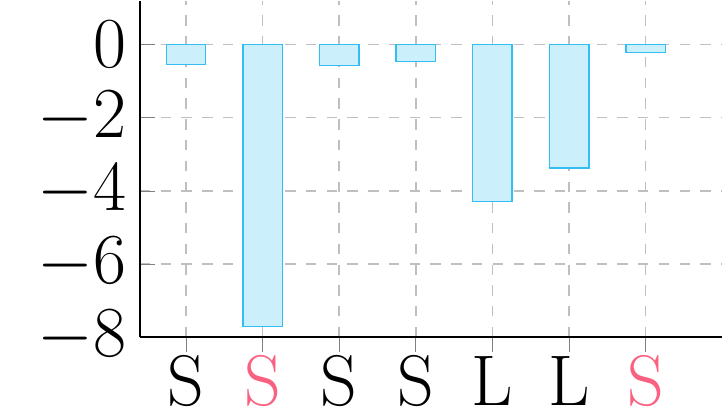}}} &
    \multicolumn{14}{c}{\multirow{4}[2]{*}{\includegraphics[scale=0.25]{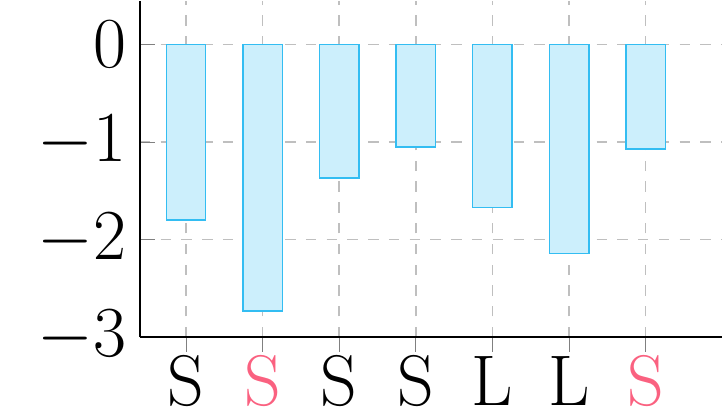}}} &
    \multicolumn{14}{c}{\multirow{4}[2]{*}{\includegraphics[scale=0.25]{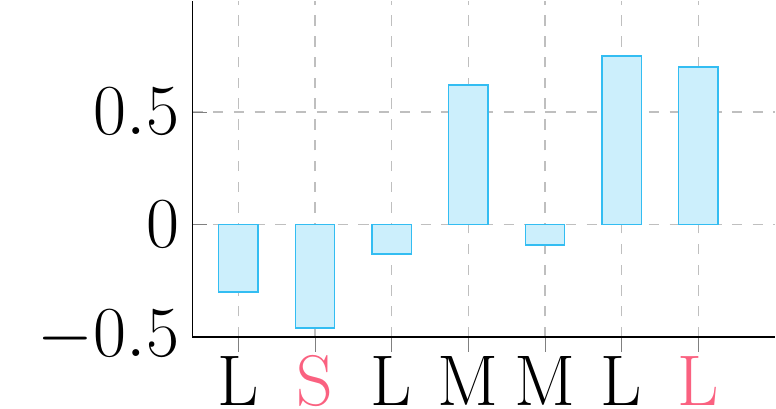}}} 
    \\
     \multicolumn{1}{l}{A: \textit{Cw2vBavgLstmCrf}} &  
    \\ 
    \multicolumn{1}{l}{B: \textit{Cw2vBavgCnnCrf}} & 
     \\  \\ 
    \multicolumn{1}{c}{\textcolor{brinkpink}{Aided-diagnosis}} & 
    \\ \\
    
     \midrule
     \multicolumn{1}{l}{Overall F1} & 
     \multicolumn{14}{c}{A:96.46; B:96.41} &
     \multicolumn{14}{c}{A:94.10; B:92.74} &
     \multicolumn{14}{c}{A:95.08; B:94.09} &
     \multicolumn{14}{c}{A:92.81; B:91.40}&
     \multicolumn{14}{c}{A:93.67; B:93.25}&
     \multicolumn{14}{c}{A:92.04; B:92.00}&
     \multicolumn{14}{c}{A:94.71; B:93.16} \\
    \cmidrule(r){1-1}\cmidrule(lr){2-15}\cmidrule(lr){16-29}\cmidrule(lr){30-43}\cmidrule(lr){44-57}\cmidrule(lr){58-71}\cmidrule(lr){72-85}\cmidrule(lr){86-99}
    
  & \multicolumn{14}{c}{\multirow{3}[2]{*}{\includegraphics[scale=0.25]{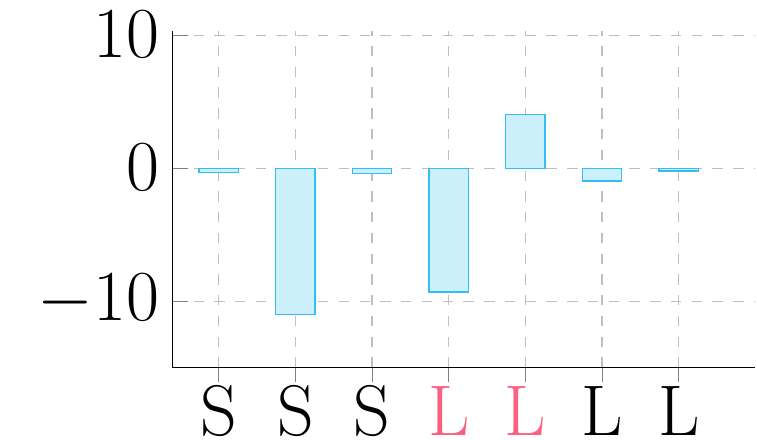}}}              & \multicolumn{14}{c}{\multirow{3}[2]{*}{\includegraphics[scale=0.25]{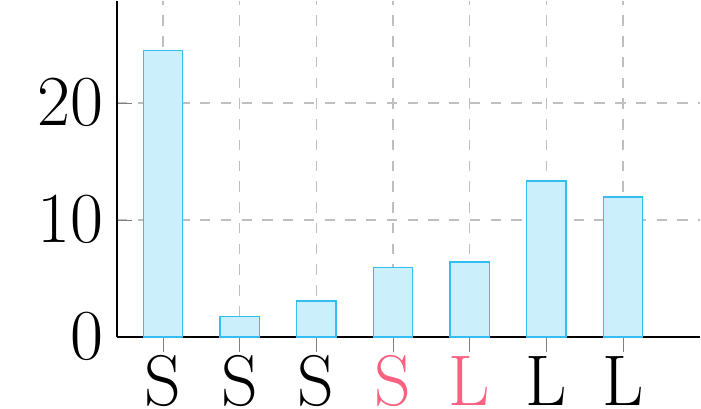}}}              & \multicolumn{14}{c}{\multirow{3}[2]{*}{\includegraphics[scale=0.25]{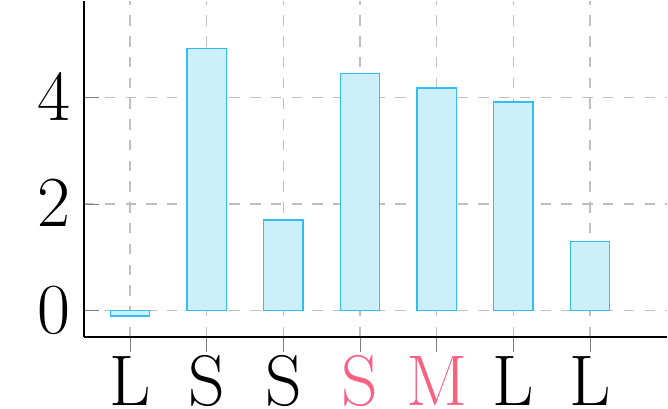}}}              & \multicolumn{14}{c}{\multirow{3}[2]{*}{\includegraphics[scale=0.25]{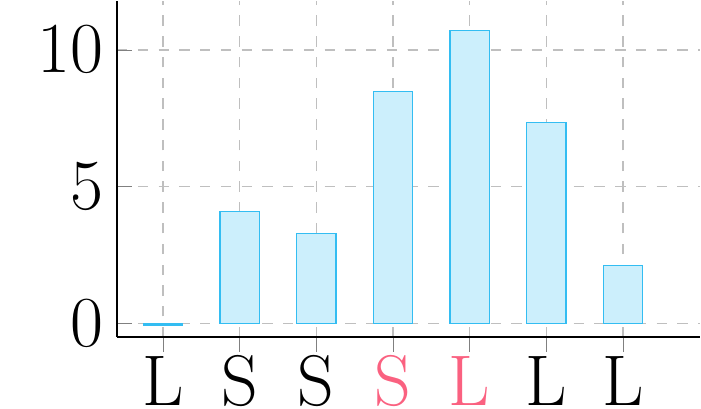}}} &
    \multicolumn{14}{c}{\multirow{3}[2]{*}{\includegraphics[scale=0.25]{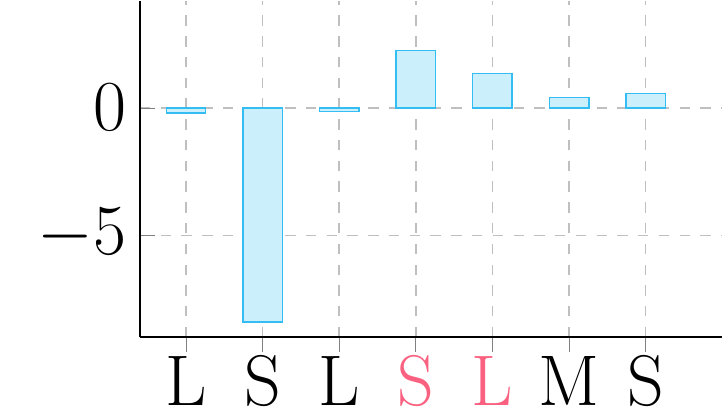}}} &
    \multicolumn{14}{c}{\multirow{3}[2]{*}{\includegraphics[scale=0.25]{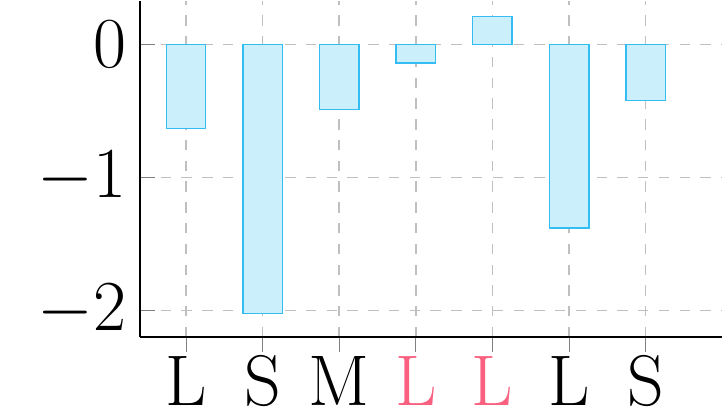}}} &
    \multicolumn{14}{c}{\multirow{3}[2]{*}{\includegraphics[scale=0.25]{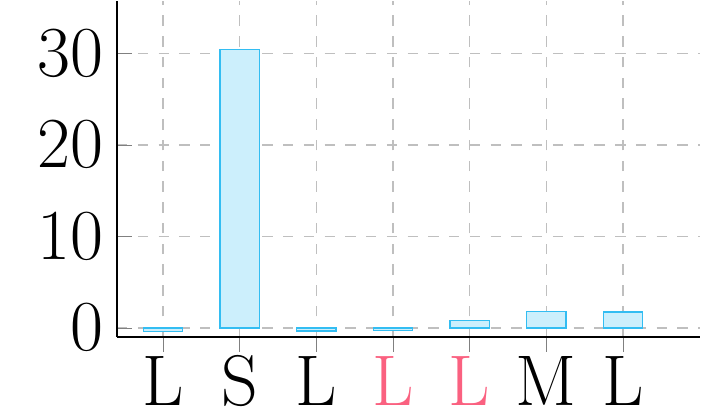}}} 
    \\
     \multicolumn{1}{l}{A: \textit{Cw2vBavgLstmCrf}} &  
    \\ 
    \multicolumn{1}{l}{B: \textit{Cw2vBavgLstmMlp}} &
     \\  \\ 
    \multicolumn{1}{c}{\textcolor{brinkpink}{Aided-diagnosis}} & 
    \\ \\

     \midrule
     \multicolumn{1}{l}{Overall F1} & 
     \multicolumn{14}{c}{A:96.46; B:96.21} &
     \multicolumn{14}{c}{A:94.10; B:94.22} &
     \multicolumn{14}{c}{A:95.08; B:95.32} &
     \multicolumn{14}{c}{A:92.81; B:92.81}&
     \multicolumn{14}{c}{A:93.67; B:93.54}&
     \multicolumn{14}{c}{A:92.04; B:92.01}&
     \multicolumn{14}{c}{A:94.71; B:94.87} \\
      \cmidrule(r){1-1}\cmidrule(lr){2-15}\cmidrule(lr){16-29}\cmidrule(lr){30-43}\cmidrule(lr){44-57}\cmidrule(lr){58-71}\cmidrule(lr){72-85}\cmidrule(lr){86-99}
      
  & \multicolumn{14}{c}{\multirow{3}[2]{*}{\includegraphics[scale=0.25]{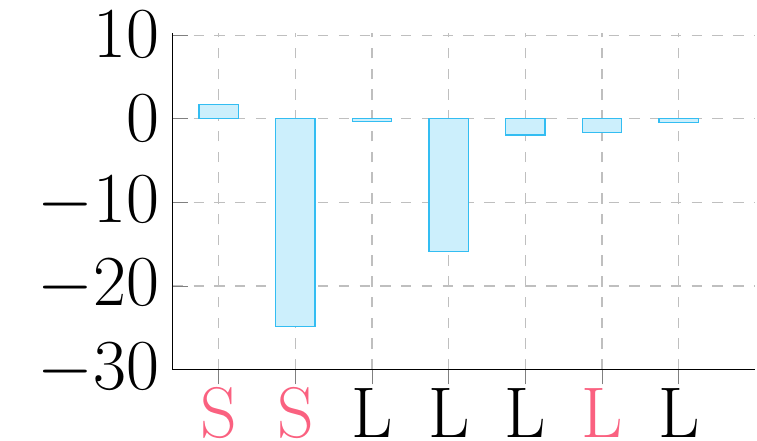}}}              & \multicolumn{14}{c}{\multirow{3}[2]{*}{\includegraphics[scale=0.25]{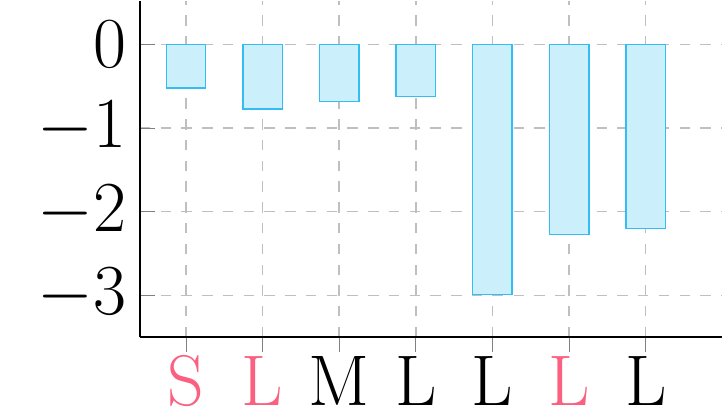}}}              & \multicolumn{14}{c}{\multirow{3}[2]{*}{\includegraphics[scale=0.25]{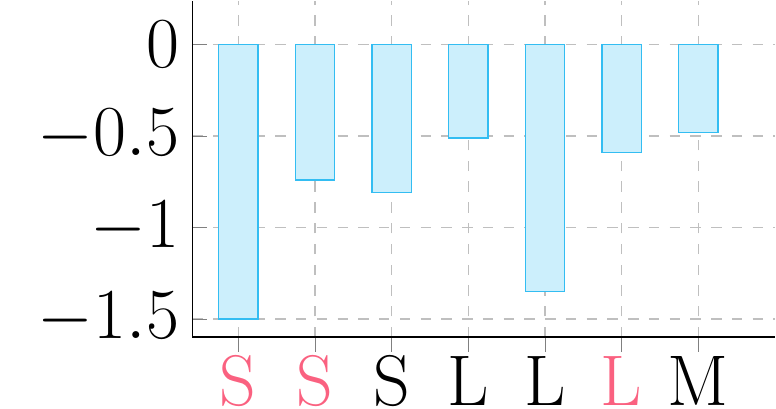}}}              & \multicolumn{14}{c}{\multirow{3}[2]{*}{\includegraphics[scale=0.25]{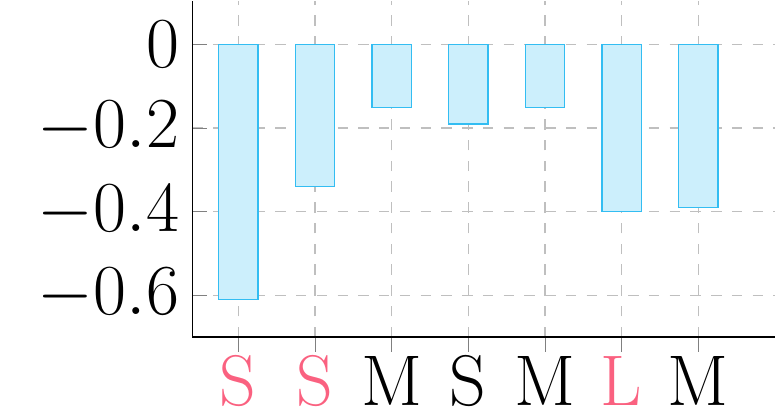}}} &
    \multicolumn{14}{c}{\multirow{3}[2]{*}{\includegraphics[scale=0.25]{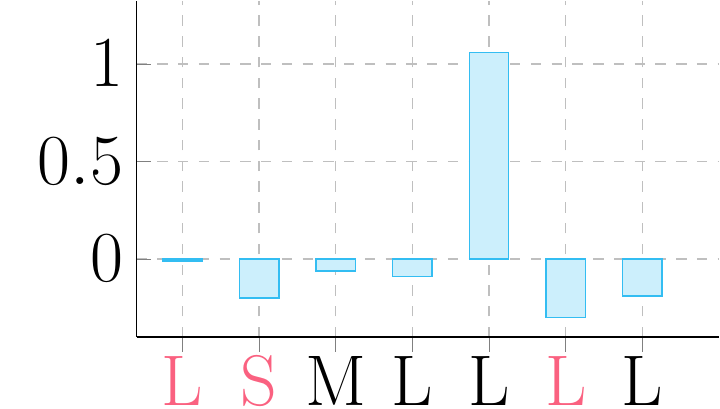}}} &
    \multicolumn{14}{c}{\multirow{3}[2]{*}{\includegraphics[scale=0.25]{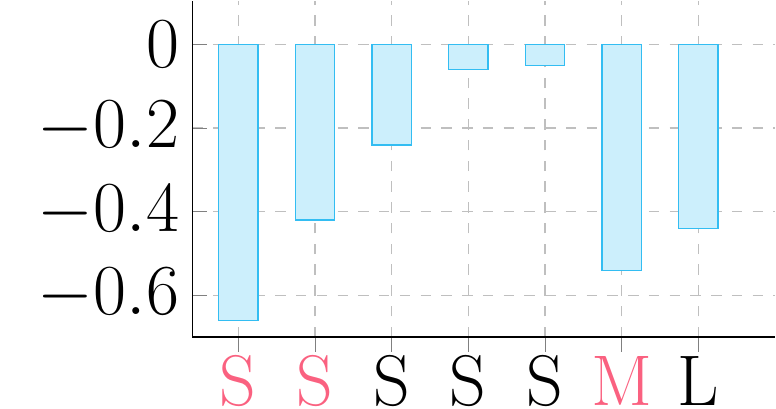}}} &
    \multicolumn{14}{c}{\multirow{3}[2]{*}{\includegraphics[scale=0.25]{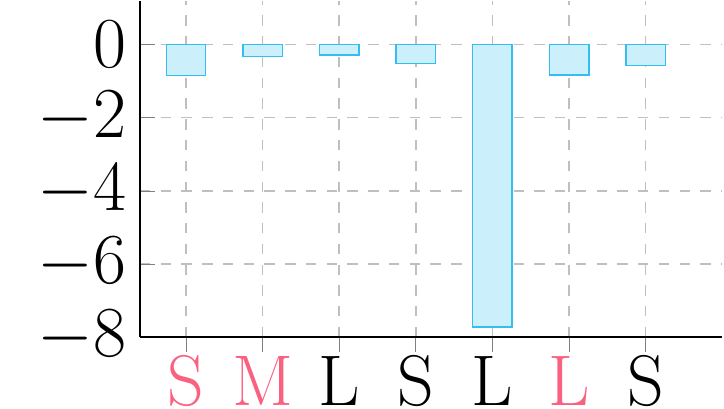}}} 
    \\
     \multicolumn{1}{l}{A: \textit{Cw2vBavgLstmCrf}} &  
    \\ 
    \multicolumn{1}{l}{B: \textit{CrandBavgLstmCrf}} & 
     \\  \\ 
    \multicolumn{1}{c}{\textcolor{brinkpink}{Aided-diagnosis}} & 
    \\ \\
    \bottomrule
    \end{tabular}%
  \caption{Diagnosis of different CWS systems. 
  For ease of presentation, we attribute values are classified into three categories: \textbf{small}(S), \textbf{middle}(M),  and \textbf{large}(L). 
  Regarding \textit{Aided-diagnosis}, the bins below the line ``$x=0$'' represent the largest gap that model $A$ is less than model $B$. By contrast, the bins above the line ``$x=0$'' denote the largest gap that model $A$ is better than model $B$.
  x-ticklabels in red indicate that the corresponding bins will be used for analysis in this section}
  \label{tab:bucket-wise-appendix}%
\end{table*}%

\section{Application: Model Diagnosis}
\label{sec:model-diagmosis-appendix}
Model diagnosis is the process of identifying where the model works well and where it worse. Tab.~\ref{tab:bucket-wise-appendix} shows several model diagnoses of different CWS systems.
Below, we will give several comparative-diagnostic analysis on some typical models.

\paragraph{LSTM v.s. CNN} 
For the choice of CNN or LSTM, the main factors are  \texttt{sLen} (sentence length) and \texttt{cCon} (label consistency of characters), referring to third row of Tab.~\ref{tab:bucket-wise-appendix}.
Besides shorter sentences, we're surprised to find that the \textit{CNN encoder is better at handling ambiguous characters than LSTM}.
Generally, we believe that LSTM could provide more long-term information, therefore, achieving disambiguation.
However, the above results show that local information is more important to learn these highly ambiguous characters 
(such as ``\texttt{的}'',``\texttt{了}'', ``\texttt{什}'') for the CWS task.
Based on this, we could explain why CNN outperforms LSTM on \texttt{ncc} (lowest value of $\alpha^{u}_{wCon}$) while is significantly worse than LSTM on \texttt{ctb} and \texttt{sxu} (large values of $\alpha^{u}_{cAmb}$).

\paragraph{CRF v.s. MLP}
CRF decoder has no advantage in dealing with unambiguous words compared with MLP, but is superior in processing long (\texttt{wLen=L})  and ambiguous (\texttt{wCon=S}) words, as observed in the fourth row of Tab.~\ref{tab:bucket-wise-appendix}.
Particularly, \textit{Cw2vBavgLstmCrf} outperforms \textit{Cw2vBavgLstmMlp} models by a large margin in the bucket of (\texttt{wCon=S}) on the \texttt{pku} dataset.
Based on this, we could explain the difference in the holistic F1 results between the above two models.

\paragraph{Cw2vBavg v.s. CrandBavg} 
As shown in the last row of the Tab.~\ref{tab:bucket-wise-appendix}, we find that pre-trained knowledge does not always help to improve the performance, especially when:
1) the characters or words are highly ambiguous;
2) the OOV density of a sentence is high.
Above evidences will help us to explain why \textit{CrandBavg} could achieve better performance measured on the holistic F1 on \texttt{ctb}, \texttt{sxu} and \texttt{pku}. They share a property of much higher value of $\alpha_{wCon}^{\mu}$, $\alpha_{cCon}^{\mu}$, $\alpha_{oDen}^{\mu}$ as observed in the Fig. 2 (a). 

\end{CJK*}

\end{document}